\documentclass{article}

\usepackage{microtype}
\usepackage{graphicx}
\usepackage{subcaption}
\usepackage{booktabs} %
\usepackage{acronym}
\usepackage{color-edits}
\usepackage{placeins}
\addauthor[Waiss]{WA}{blue}
\addauthor[Eug]{EU}{orange}

\usepackage{hyperref}

\usepackage[accepted]{icml_r2fm}

\usepackage{amsmath}
\usepackage{amssymb}
\usepackage{mathtools}
\usepackage{amsthm}
\usepackage{enumitem}

\usepackage[capitalize,noabbrev]{cleveref}

\theoremstyle{plain}

\theoremstyle{definition}

\theoremstyle{remark}

\usepackage[textsize=tiny]{todonotes}

\icmltitlerunning{The Geometries of Truth Are Orthogonal Across Tasks}

\begin{document}
\newacro{LLM}[LLM]{Large Language Model}
\twocolumn[
\icmltitle{The Geometries of Truth Are Orthogonal Across Tasks}

\begin{icmlauthorlist}
\icmlauthor{Waiss Azizian}{Apple}
\icmlauthor{Michael Kirchhof}{Apple}
\icmlauthor{Eugene Ndiaye}{Apple}
\icmlauthor{Louis Bethune}{Apple}
\icmlauthor{Michal Klein}{Apple}
\icmlauthor{Pierre Ablin}{Apple}
\icmlauthor{Marco Cuturi}{Apple}
\end{icmlauthorlist}

\icmlaffiliation{Apple}{Apple}

\icmlcorrespondingauthor{Michael Kirchhof, Marco Cuturi}{contact see website}

\icmlkeywords{Machine Learning, ICML}

\vskip 0.3in
]

\printAffiliationsAndNotice{\icmlEqualContribution} %
\begin{abstract}
Large Language Models (LLMs) have demonstrated impressive generalization capabilities across various tasks, but their claim to practical relevance is still mired by concerns on their reliability. %
Recent works have proposed examining the activations produced by an LLM at inference time to assess whether its answer to a question is correct.
Some works claim that a ``geometry of truth'' can be learned from examples, in the sense that the activations that generate correct answers can be distinguished from those leading to mistakes with a linear classifier.
In this work, we underline a limitation of these approaches: we observe that these "geometries of truth" are intrinsically task-dependent and fail to transfer across tasks.
More precisely, we show that linear classifiers trained across distinct tasks share little similarity and, when trained with sparsity-enforcing regularizers, have almost disjoint supports. 
We show that more sophisticated approaches (e.g., using mixtures of probes and tasks) fail to overcome this limitation, likely because activation vectors commonly used to classify answers form clearly separated clusters when examined across tasks. 
\end{abstract}

\section{Introduction}

Large Language Models (LLMs) have seen tremendous success in recent years across a wide range of tasks. However, their widespread deployment is not without risks: from hallucinations \citep{ji2023survey} to outright deception \citep{park2024ai}, 
the complexities underpinning LLM generation can be the root of many issues. These challenges are particularly concerning in high-stakes domains like healthcare, legal advice, and financial analysis, where incorrect or misleading information can lead to serious harm.
As a result, several works have suggested leveraging the various activations generated by a model at inference time to understand and assess the truthfulness of its output. 
\citet{azaria2023internal} demonstrated that training a simple classifier on top of the hidden activations of {LLMs} can help predict whether an LLM has provided a truthful answer to a user-provided question. This finding suggests that models somewhat reveal enough information in their activations at inference time to help users assess whether they are producing correct information. %
Numerous subsequent works have explored this ``geometry of truth'' and confirmed that it is approximately linear, in the sense that a linear classifier can distinguish reliably truthful from erroneous answers~\citep{li2023inference,marks2024the,xiong2024efficient,burns2023discovering,kossen2025semantic}.
As a token of their relevance, these directions can then be used to steer the {LLM} towards factual generations \citep{li2023inference,wang2024adaptive}, and are increasingly studied as a cheap and effective proxy to assess the uncertainty of an {LLM} output on a given task, e.g.~\citep{sky2024androids,zhang2025reasoningmodelsknowtheyre,slobodkin2023curious}. %

\textbf{But is one hyperplane all it takes?} While the literature abounds with examples of the relevance of such probes when tested within a given task or knowledge domain, it remains unclear whether they can generalize across different tasks. Although some works have reported encouraging results in that direction \citep{azaria2023internal, marks2024the, beigi2024internalinspector}, others provide a mixed assessment \citep{slobodkin2023curious, kossen2025semantic, zhang2025reasoningmodelsknowtheyre}, while some works \citep{orgad2025llms, levinstein2024still, sky2024androids} show on the contrary that probes completely fail to generalize in some settings. In that context, one might be tempted to increase the complexity of probes and their training procedures, such as the more sophisticated pipeline proposed by \citet{beigi2024internalinspector} that incorporates augmentations to build additional representations on top of the activations of multiple layers taken jointly. Taken together, these works paint a mixed picture of whether simple probes can generalize at all, and if they do not, whether their disappointing results are an artifact of their training or an intrinsic limitation. 

\textbf{Our contributions.} Our work proposes to answer whether there is any hope to see linear probes transfer reliably \emph{across tasks}.
We study the variability of these probes across tasks and reach the following findings:

\textbf{Task-specific truthfulness geometries.} We first demonstrate that truthfulness geometries are fundamentally task-specific. Through comprehensive cross-task evaluation on seven diverse datasets, we show that linear probes trained on different tasks exhibit distinct "geometries of truth" that fail to generalize. While some task pairs show successful transfer, most combinations result in substantial performance degradation. This systematic analysis shows that generalization success depends critically on task similarity rather than being a universal property of truthfulness probes.

\textbf{Geometric analysis of orthogonality.} We then provide geometric analysis revealing why these failures occur. We demonstrate that truthfulness directions are largely orthogonal across tasks, with a clear correlation between geometric similarity and generalization performance. Using sparse probes, we reveal that probe supports are nearly disjoint across tasks, providing interpretable evidence for orthogonality. Visualizations show that different tasks form distinct clusters in the representation space of the model, confirming our geometric explanation for the failure to transfer.

\textbf{Orthogonality persists in multi-task settings.} Finally, we test whether our orthogonality hypothesis holds when training on mixtures of task. We demonstrate that training on diverse task mixtures fails to resolve generalization problems, and critically show that optimal directions for target tasks cannot be recovered through linear combinations of directions from other tasks. We further show that more complex architectures are also no better than naive parameter summation, suggesting the limitation is intrinsic. Given these findings, we explore conservative deployment strategies using conformal prediction to maintain reliability guarantees in cross-task scenarios.

Our work is structured as follows:
\begin{itemize}[leftmargin=.3cm,itemsep=.05cm,topsep=0cm,parsep=2pt]
\item In \cref{sec:bg}, we review the necessary background on probing and present a detailed survey of the literature on generalization properties of uncertainty probes.
\item In \cref{sec:task-to-task}, we introduce our experimental setup (models and datasets) and systematically study cross-task generalization failures through geometric analysis of probe directions and sparse probe supports.
\item In \cref{sec:mt}, we examine whether training on mixtures of tasks can overcome these limitations, test more complex architectures, and explore conservative deployment strategies.
\end{itemize}

\section{Background}
\label{sec:bg}

\paragraph{Linear probing for uncertainty quantification.}
Uncertainty quantification for LLMs has attracted significant attention recently. %
Apart from relatively more costly multi-samples methods such as semantic entropy \citep{farquhar2024detecting}, many efforts have focused on learning simple classifiers whose inputs are activation vectors computed by the LLM at inference time, using labeled pairs of questions and either correct or incorrect answers.
\citet{azaria2023internal} introduced this approach for uncertainty quantification for LLMs. While the classifier was originally set to be a multi-layer perceptron, follow-up works showed that using a simple linear logistic regression model could achieve similar performance \citep{li2023inference,orgad2025llms,marks2024the,santilli2025revisiting}. In other words, and following \citeauthor{marks2024the}'s terminology, there might be a linear ``geometry of truth'' that can separates the representations of correct from incorrect outputs.

\paragraph{The Geometry of Truth Hypothesis.}
Given a user question $q$, the LLM generates an answer $\hat a = (\hat a_1, \ldots, \hat a_T)$ autoregressively. At each step $t \in [T]$, the model produces hidden state vectors $h_{t, \ell} \in \mathbb{R}^d$ where $\ell$ indexes the layer and $t$ the token position. In this work, we focus on a fixed layer $\ell^\star$ (e.g. 28 and 21 for Qwen models) and extract the representation $h_{T, \ell^\star}$ at the final token of the output (or the token before, $T-1$).
To provide supervision, we follow \citep{farquhar2024detecting,santilli2025revisiting} and label the correctness of $\hat a$ given the gold answer $a$ using LLM-as-a-judge \citep{zheng2023judging}. This yields a label $y \in \{-1,+1\}$ which is positive is the answer $\hat a$ is correct and negative otherwise.
This provides a dataset $\mathcal{D} = \{(h_i, y_i)\}_{i\in [N]}$ where each point $h_i = h_{T, \ell^\star}$ is the final hidden representation for a generated answer $i$. It is the internal state right before the model decides what the final answer token will be. We then train a linear probe i.e. a logistic regression classifier to predict correctness from the hidden states:
\begin{equation}
\min_{\substack{\theta \in \mathbb{R}^d,\\b \in \mathbb{R}}}
\frac1N \sum_{i=1}^{N} \log\left(1 + e^{-y_i(\theta^\top h_i + b)}\right) + \frac{\lambda_2}{2} \|\theta\|_2^2
\label{eq:logistic_regression}
\end{equation}
The \emph{geometry of truth hypothesis} states that truthful and untruthful generations are linearly separable in model's hidden space, such that a single hyperplane parametrized by $(\theta, b)$ can distinguish correct from incorrect outputs. We explore if this is universal or task-specific.

\paragraph{On the Generalization of Uncertainty Probes.}
The generalization of uncertainty estimates has been studied in several works but there is still no definitive consensus, neither among papers nor within the papers themselves, as many of them conclude with a nuanced assessment. Some of the historically first results were promising:
\begin{itemize}[leftmargin=.3cm,itemsep=.05cm,topsep=0cm,parsep=2pt]
\item \citet{azaria2023internal} introduce their own dataset of facts, the \textit{true-false} dataset, made of several splits with different subjects. With their carefully constructed dataset, they show that probes trained on several subjects can accurately predict the correctness of facts of another subject.
\item \citet{kapoor2024large} consider finetuning the whole LLM to obtain better uncertainty probes;
although they do not conduct systematic evaluations, they report good performance on a couple of datasets that were not seen during training.
\end{itemize}

Other works apply probes in slightly different contexts and report mixed generalization results. 
\begin{itemize}[leftmargin=.3cm,itemsep=.05cm,topsep=0cm,parsep=2pt]
\item \citet{slobodkin2023curious} leverage probes to classify unanswerable queries and study their generalization performance on datasets not seen at training (their Figure 6). Although performance decreases, the authors still note that it remains better than probes trained on the first hidden layer on the correct dataset. 
\item \citet{zhang2025reasoningmodelsknowtheyre} use uncertainty probes for reasoning tasks and find that, although these probes generalize across similar datasets, they fail when the type of reasoning required changes. 
\item \citet{kossen2025semantic} suggest training probes with semantic entropy \citep{farquhar2024detecting} and show that, for generalization to new tasks, this improves robustness in the choice of layer (their Figure 6).
\item \citet{kadavath2022language} introduces a finetuning approach trained on a diverse mixture of tasks which is evaluated both on a held-out dataset and when training on only one task: the authors observe a decrease in performance but still note decent generalization.
\end{itemize}

Then there are paper that report generally negative generalization results.
\begin{itemize}[leftmargin=.3cm,itemsep=.05cm,topsep=0cm,parsep=2pt]
\item \citet{beigi2024internalinspector} report significantly worse performance on out-of-distribution data (Table 3) despite using all hidden states of an {LLM} as an input for an uncertainty estimator.
\item \citet{marks2024the} study, in a controlled setting similar to \citep{azaria2023internal}, the geometry of representations of correct and incorrect answers as well as generalization properties of probes across datasets. Though the authors note that training on datasets and their negations helps, the performance of probes on out-of-distribution data remains suboptimal (Figure 5). They also visualize the hidden spaces at different layers to understand when a linear representation of uncertainty emerges.
\item \citet{levinstein2024still} reproduce the setting of \citet{azaria2023internal} and note that their probes catastrophically fail to generalize under trivial changes like introducing negations (\S 4.4), with a roughly 20\% accuracy loss.
\item In  \citet{orgad2025llms}, the authors take a systematic approach: they consider a wide variety of question-answers datasets and find that probes trained on each of these datasets fail to generalize (their Figure 3).
\item \citet{sky2024androids} consider ensembles of attention-based probes for hallucination detection but notes that they do not generalize, both when trained on one dataset and tested in another (their Table 6) and when trained in two tasks and tested in a third (Table 8). This work raises the question of whether having a larger mixture of tasks night help, which we will answer by the negative in this work. %
\end{itemize}

\section{Truthfulness directions across tasks}
After having reviewed the necessary background and the literature on the generalization of probes, we now present our experimental setup before starting our study of cross-task generalization.

\textbf{Training Probes.}
To obtain uncertainty estimates with linear probes, we follow standard practices \citep{li2023inference,orgad2025llms,marks2024the,santilli2025revisiting} and train the probes using the logistic regression implementation of \citet{scikit-learn}. These probes are trained with $L_2$ regularization with the hyperparameter tuned with cross-validation on the training set.
We consider two standard token positions $t$:  the stop token of the output and the token before the stop token of the output. Similarly, we base the probes on two embeddings: those of the last hidden layer and those at 75\% depth. 

\textbf{Models.}
We consider the following models: Qwen 2.5 7B Instruct \cite{qwen2024qwen205}, Phi 4 Mini \citep{abdin2024phi04} and Llama 3.1 8B Instruct \citep{grattafiori2024llama}.
In the main text, we consider probes trained via \cref{eq:logistic_regression} that operate on the stop token at the last layer (layer 28) of Qwen 2.5 7B Instruct. All other combinations give similar results, which we report in the appendix.

\paragraph{Datasets.}
To study generalization, we use several datasets that are variously related: 
NQ \citep{kwiatkowski2019natural} and SimpleQA \citep{wei2024measuring} are general question-answering datasets. TriviaQA \citep{joshi2017triviaqa}, and
SQUAD \citep{rajpurkar2016squad} are reading-comprehension datasets with a context, but also on generic topics. BioASQ \citep{nentidis2023bioasq} is composed of  biology questions.
SVAMP \citep{patel2023svamp}, made of simple arithmetic questions, and GSM8K \citep{cobbe2021gsm8k} of more complex math word problems.

\label{sec:task-to-task}

\begin{figure}[h]
    \centering
    \begin{subfigure}[t]{\linewidth}
    \includegraphics[width=\linewidth,trim={0.cm 0.cm 0cm 0.2cm},clip]{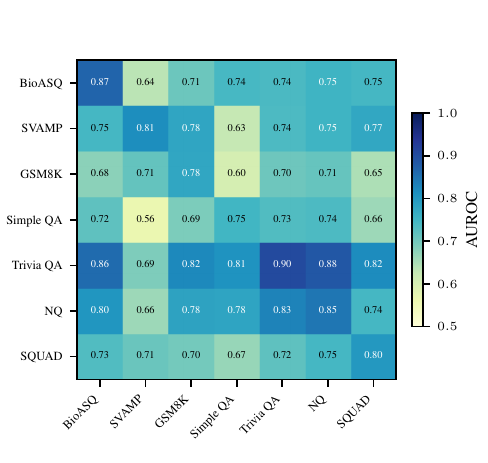}
    \end{subfigure}
    \begin{subfigure}[t]{\linewidth}
    \includegraphics[width=\linewidth,trim={0.cm 0.cm 0.cm 0.2cm},clip]{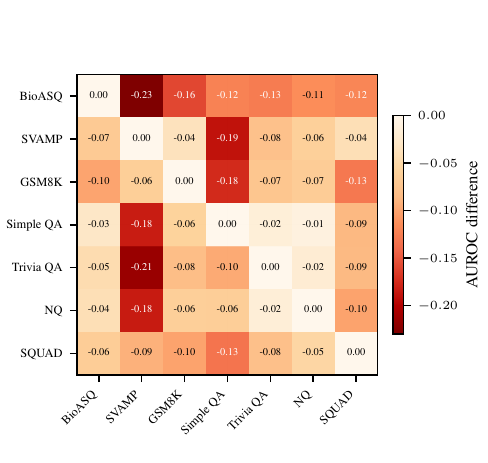}
    \end{subfigure}
    \caption{AUROC of probes trained on different tasks on the stop token of the output on last layer. Rows correspond to evaluation tasks while columns correspond to training tasks. The second plot represents the difference between the probe trained on this task and probes trained on the other datasets. Results are averaged over 5 runs.}
    \label{fig:pairwise_auroc}
\end{figure}

\paragraph{Cross-task generalization failures.}
We train linear probes for each dataset, then systematically evaluate cross-task transfer performance. Our dataset collection enables analysis of both successful and failed generalization cases.
\Cref{fig:pairwise_auroc} confirms widespread generalization failures, in line with \citet{orgad2025llms}, but reveals a crucial pattern: tasks cluster into semantically coherent groups. Factual recall tasks (TriviaQA, NQ, SimpleQA) show mutual transferability, while specialized domains (BioASQ biology, GSM8K/SVAMP mathematics) remain isolated. We thus confirm that probes fail to generalize to new tasks in general but also that the situation is more nuanced. Indeed, this suggests that truthfulness representations adapt to domain-specific reasoning patterns rather than capturing one universal truth signal.

\begin{figure}[h]
    \centering
    \includegraphics[width=\columnwidth]{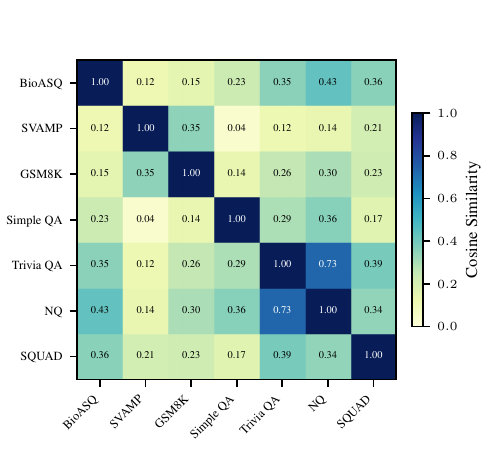}
    \caption{Cosine similarity between probes trained on different datasets using L2 regularization. Results are averaged over 5 runs.
    The cosine similarity between probe directions is consistently low (less than 0.5). Task pairs with large similarity Trivia QA - NQ (above 0.7) are the ones showing good generalization.
    }
    \label{fig:cosine}
\end{figure}

\begin{figure}[h]
    \centering
    \includegraphics[width=\columnwidth, clip, trim={0cm 0cm 0cm 1cm}]{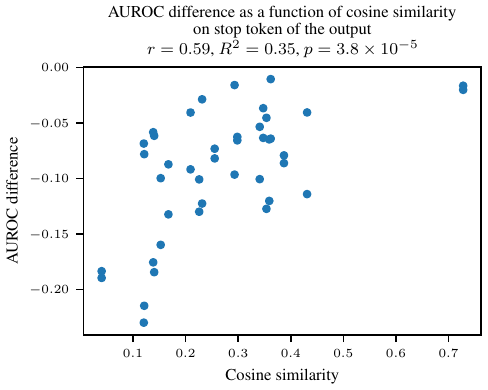}
    \caption{AUROC difference to probe trained on the right dataset as a function of cosine similarity between probes ($r = 0.59$, $R^2= 0.35$, $p = 3.8 \times 10^{-5}$)}
    \label{fig:auroc_cosine}
\end{figure}

\paragraph{Underlying geometry.}
While previous work has documented the poor cross-task generalization of truthfulness probes, the geometric basis for this failure remains unexplored. We test whether generalization patterns reflect fundamental differences in how truthfulness is encoded across task domains, rather than mere statistical artifacts of limited training data.
We examine cosine similarities between probe weight vectors (\cref{fig:cosine}). Most probe pairs exhibit near-orthogonal directions (cosine similarity $< 0.5$), with successful transfer occurring only between geometrically aligned probes. Indeed, \Cref{fig:auroc_cosine} shows that more similar probes also have better generalization performance in each others task (correlation coefficient of 0.59), showing that this is not just a mere geometrical problem but the cause for the performance drops; similar patterns with even lower cosine similarities are shown in the appendix for different models.
These findings challenge the notion of a universal "geometry of truth." Instead, truthfulness directions emerge from task-specific representational structures, with different domains occupying orthogonal subspaces in the model's hidden state space. However, cosine similarity provides only a coarse measure of geometric relationships. To gain deeper insight into which specific features drive these differences, we next examine the support structure of sparse probes.

\begin{figure}[h]
    \centering
    \includegraphics[width=\columnwidth]{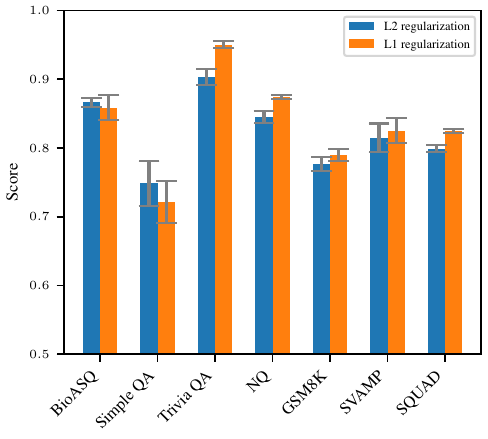}
    \caption{AUROC of linear probes with L1 or L2 regularisation. Results are averaged over 5 runs.}
    \label{fig:comparison_l1_l2}
\end{figure}

\begin{figure}[h]
    \centering
    \includegraphics[width=\columnwidth]{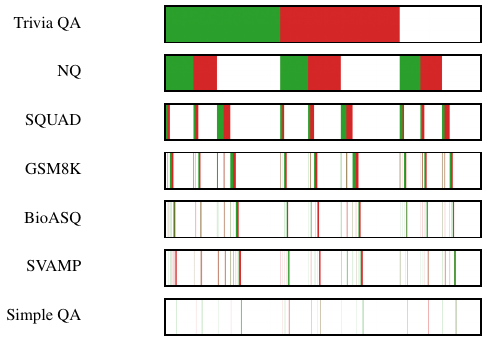}
    \caption{Signed support of sparse probes trained on different datasets at the stop token of the output, using L1 regularization at layer 28. Each row represents one probe trained on the corresponding dataset (y-axis labels). The x-axis shows the 3584 dimensions of the hidden state vector. Green indicates positive coefficients, red indicates negative coefficients, and white indicates zero coefficients (sparsity). Dimensions are sorted by sparsity level across all datasets, with the least sparse dimensions on the left and the most sparse on the right.}
    \label{fig:supports}
\end{figure}

\begin{figure}[h]
    \centering
    \includegraphics[width=\columnwidth]{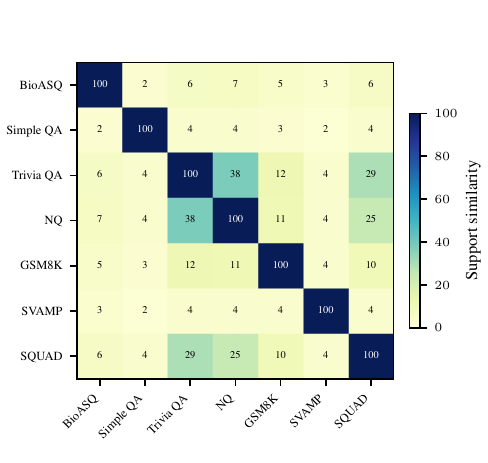}
    \caption{Support overlap between sparse probes trained on different datasets. Darker colors indicate higher overlap percentages. Task pairs with $>30\%$ overlap (TriviaQA, NQ) correspond to successful cross-task generalization, while most pairs show $<15\%$ overlap, explaining generalization failure.}

    \label{fig:support_percentages}
\end{figure}

\label{sec:sparse}
\paragraph{Sparse probes.}
To deepen our understanding of the features learned by probes, we consider sparse probes obtained through \textit{sparse} logistic regression. Instead of \cref{eq:logistic_regression}, we consider probes trained with $L_1$ regularization:
\begin{equation}
\min_{\substack{\theta \in \mathbb{R}^d,\\b \in \mathbb{R}}} \frac1N \sum_{i=1}^{N} \log\left(1 + e^{-y_i(\theta^\top h_i + b)}\right)
 + {\lambda_1} \|\theta\|_1\,.
    \label{eq:sparse_logistic_regression}
\end{equation}
 The $\ell_1$-regularization coefficient $\lambda_1$ is tuned on a held-out validation set.
As shown in \cref{fig:comparison_l1_l2}, this approach does not degrade performance. But it provides us with a visual way of comparing probes in \cref{fig:supports}: The supports of probes between tasks are nearly disjoint. This can be made formal by computing support overlap, see \cref{fig:support_percentages}. We observe the same pattern as in \cref{fig:cosine}: some tasks have more support overlap than others and this coincides with the ones where probes generalize better.

\paragraph{Task-specific geometries.}

\begin{figure}[h]
    \centering
    \begin{subfigure}[t]{\columnwidth}
    \includegraphics[width=\textwidth]{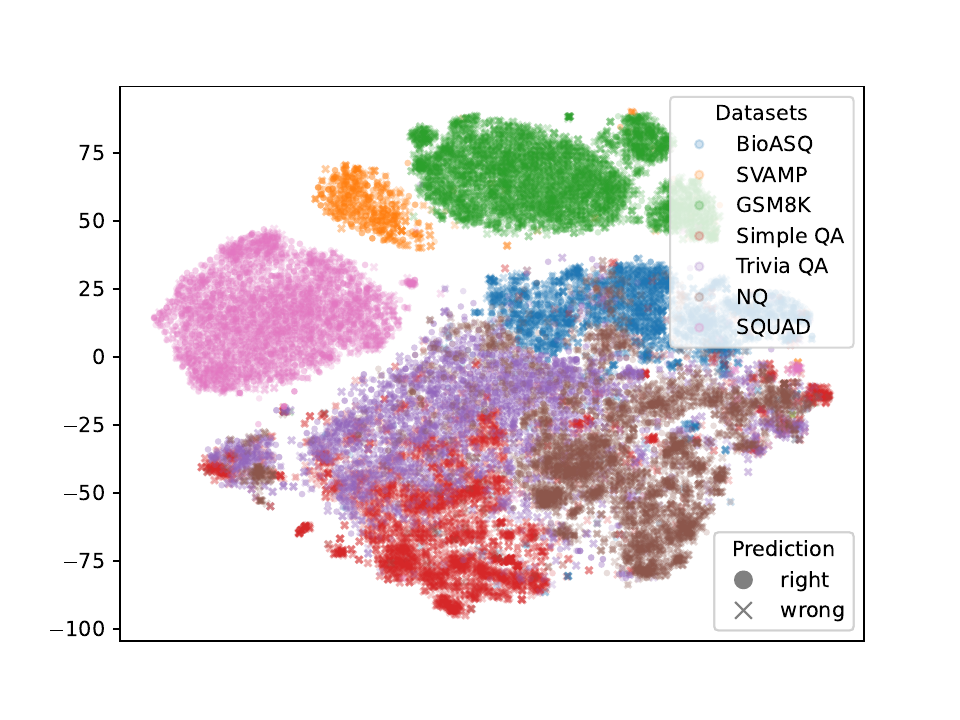}
    \end{subfigure}
    \caption{t-SNE plots of the hidden space at layer 28 at the stop token of the output. Different tasks form distinct clusters in representation space, with the correct/incorrect distinction being secondary to task boundaries.}
    \label{tsne}
\end{figure}

The previous experiments indicate that geometric information and generalization (failures) align with semantic task properties. %
This suggests that truthfulness detection mechanisms are not universal but rather emerge from task-specific representations within the model's hidden states. To visualize this phenomenon, we provide t-SNE plots \citep{van2008visualizing} of the hidden states of the model for each task \cref{tsne}.
What we see is that most of the tasks actually form distinct clusters. Similar tasks such as TriviaQA, NQ, SimpleQA are slightly mixed, but BioASQ or mathematical reasoning tasks are clearly separated from each other. The uncertainty signal, represented by the two right or wrong classes, is secondary compared to the distinction between tasks.
This confirms again our hypothesis that the truthfulness information is not universal and is very much task-dependent.

\section{Generalization from mixture of tasks}
\label{sec:mt}

\subsection{Orthogonality in mixture of tasks}

We now revisit our orthogonality hypothesis in a multi-task setting. We consider learning probes not on only one dataset but on a mixture of them, examining whether the geometric relationships we observed persist in this more complex training scenario.
\begin{figure*}[!h]
    \centering
    \includegraphics[width=\linewidth]{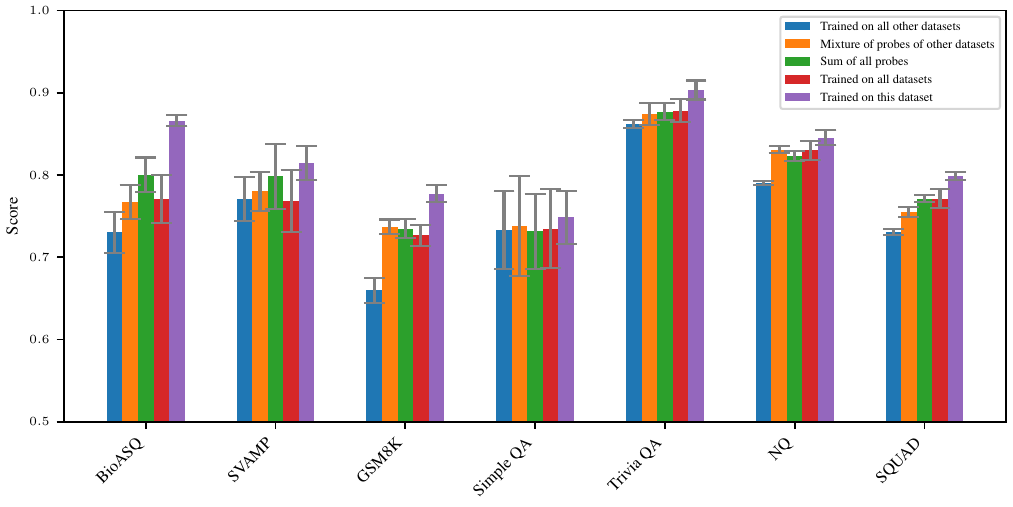}
    \caption{AUROC of linear probes at the stop token of the output in the multi-task setting, using L2 regularization.}
    \label{fig:generalization_mt}
\end{figure*}

\Cref{fig:generalization_mt} presents our main findings across several training strategies. The purple bars show the best performing probes: task-specific probes trained and tested on the same domain. The blue bars test our central question by showing performance when probes are trained on all tasks except the target task. Despite this diverse training mixture, we observe substantial performance drops across all domains, suggesting that multi-task training does not resolve the generalization problem. 

To test whether this failure stems from our orthogonality hypothesis, we examine whether truthfulness directions from other tasks can be linearly combined to recover the direction from the target task. The orange bars show results from a constrained optimization where we restrict probe coefficients to lie within the subspace spanned by probes from other tasks. Formally, given probes $\theta_1,\dots,\theta_6$ from non-target tasks, we solve:
\begin{align*}
    &\min_{\substack{\alpha \in \mathbb{R}^6\\\,b \in \mathbb{R}}}   \frac{1}{N} \sum_{i=1}^{N} \log\left(1 + e^{-y_i(\theta_\alpha^\top h_i + b)}\right)
 + \frac{\lambda_2}{2} \|\alpha\|_2 \\
    & \text{ where } \theta_\alpha = \sum_{i = 1}^6 \alpha_i \theta_i\,.
\end{align*}

The suboptimal performance of this constrained approach compared to task-specific probes confirms that target task directions lie outside the subspace generated by directions of the other tasks, providing direct evidence for our orthogonality claim.

The remaining comparisons in \Cref{fig:generalization_mt} further support this interpretation. Training on all datasets simultaneously (red bars) yields performance roughly equivalent to simply summing individually trained probe parameters (green bars). This equivalence is striking: if probe directions overlapped significantly, naive parameter summation would cause destructive interference and degrade performance. Instead, the similar results confirm that probe directions are approximately orthogonal across tasks. However, both approaches still fall slightly short of task-specific training (purple bars), demonstrating that even when all tasks are included in training, the resulting probes cannot match the performance of domain-specific optimization.

\subsection{Mixture of probes}
Given the failure of linear probes to generalize across tasks, we test whether more complex probe architectures can improve cross-task transfer performance.
We experiment with a "mixture of probes" approach inspired by Mixture-of-Experts architectures. %
Our approach uses a single gating layer with 16 expert probes, where each expert consists of a 2-layer feedforward network that takes the LLM's hidden states as input. The gating mechanism learns to route different inputs to different expert probes based on the hidden state representations.

For each target task, we train this mixture of probes on the six other tasks while using a mixture of these six tasks as a validation set. We perform grid search over hyperparameters including learning rate, weight decay, and auxiliary loss coefficients for the gating mechanism taking inspiration from \citet{fedus2021switch}.\looseness=-1

\begin{figure}[h]
    \centering
    \includegraphics[width=\columnwidth]{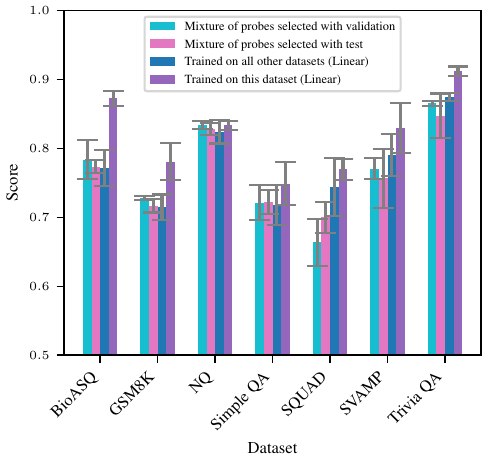}
\caption{AUROC of different methods in the multi-task setting. Mixture of probes are trained on six non-target tasks and evaluated on the target task. "Validation" uses hyperparameters selected on a validation set from the training tasks; "Test" uses hyperparameters selected on the target task (oracle setting). Linear probe baselines ("Trained on all other datasets" and "Trained on this dataset") reproduce results from \cref{fig:generalization_mt} for comparison. Results averaged over 3 runs.}
    \label{fig:moe}
\end{figure}

\Cref{fig:moe} presents results under two scenarios: an "oracle" setting where hyperparameters are selected using the test task performance, and a realistic setting where hyperparameters are chosen based on validation task performance. In both cases, performance remains below that of linear probes trained directly on the target task. Moreover, the performance of this non-linear model matches that of simple linear probes trained on the same six held-out tasks.\looseness=-1

These results show that even sophisticated probe architectures cannot bridge the performance gap with task-specific probes. The equivalence between complex and simple models trained on identical data suggests that the generalization failure stems from the orthogonal task geometries we identified, rather than limitations in model architecture.

\subsection{Conservative approaches via conformal prediction}

\begin{figure}[h]
    \centering
    \hfill
    \begin{subfigure}[t]{0.5\textwidth}
        \centering
        \includegraphics[width=\textwidth]{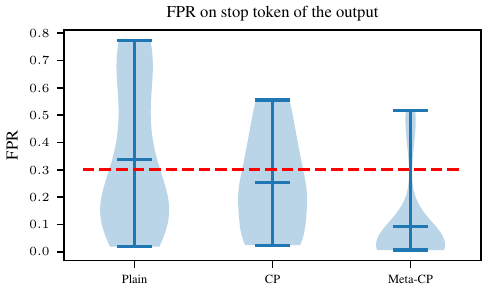}
        \caption{Violin plot of the false positive rate (FPR) of the different methods with threshold $\alpha = 0.3$.}
    \end{subfigure}
    \hfill
    \begin{subfigure}[t]{0.5\textwidth}
        \centering
        \includegraphics[width=\textwidth]{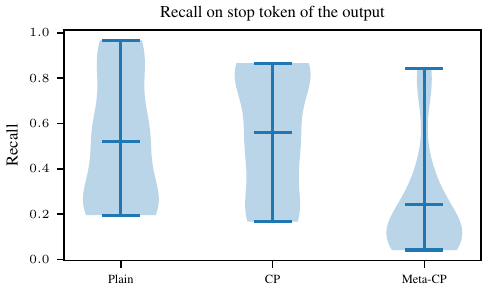}
        \caption{Violin plot of the recall of the different methods with threshold $\alpha = 0.3$.}
    \end{subfigure}
    \caption{False positive rates (FPR) and recall for thresholds tuned using different methods: standard training (Plain), split conformal prediction (CP), conformal prediction for multi-task settings (Meta-CP). The results are averaged over 5 repetitions and test tasks.}
    \label{fig:CP}
    \hfill
\end{figure}
\begin{figure}
\centering
\hfill
\begin{tabular}{lrrr}
\toprule
 Method            &   Mean FPR &   Q-80\% FPR &   Mean Recall \\
\midrule
 Plain             &       0.34 &        0.69 &          0.52 \\
 CP      &       0.25 &        0.47 &          0.56 \\
 Meta CP &       0.09 &        0.23 &          0.24 \\
\bottomrule
\end{tabular}
\caption{False positive rates (FPR) and recall for thresholds tuned using different methods: standard training (Plain), split conformal prediction (CP), conformal prediction for multi-task settings (Meta-CP). The results are averaged over 5 repetitions and means and 80\% quantiles (Q-80\%) are considered over test tasks.}
\label{tab:results}
\end{figure}

The previous sections have demonstrated that truthfulness representations are inherently task-dependent and fail to generalize across domains. This poses a critical challenge for real-world deployment: how can we maintain reliable uncertainty estimates when the distribution of user queries may differ from training data? 

Given that guaranteeing generalization appears unattainable, we explore whether conservative calibration methods can provide reliability guarantees despite poor cross-task transfer. We focus on conformal prediction as a principled approach to control error rates, examining scenarios where avoiding false endorsement of incorrect information is crucial.

We consider the setting of \cref{sec:mt} where probes are trained on multiple tasks and are evaluated on a new, unseen domain. Specifically, we train probes on all datasets except one test task using \cref{eq:logistic_regression} and seek to ensure that the false positive rate remains below a threshold $\alpha = 0.3$ on the held-out task. We compare three approaches: plain probes with default thresholds (Plain), standard split conformal prediction \citep{vovk2005algorithmic} (CP), and a variant designed for multi-task settings \citep{park2022pac} (Meta-CP).

Given a trained probe $f(h) = \theta^\top h + b$ obtained by \cref{eq:logistic_regression}, these methods calibrate a threshold $\tau$ such that the probe's confidence score must exceed $\tau$ before predicting an answer as correct. For new hidden states $h_{t,\ell}$ corresponding to question-answer $(q, a)$ with label $y$, the following bound on the false positive rate holds:
$$
\mathbb{P}
\left(
  f(h_{t, \ell})
  >
  \tau
  \,|\,
  y = -1
\right)
\leq
\alpha\,.
$$

This ensures that, on average, the false positive rate will be at most $\alpha$ for new questions. We refer to \citet{vovk2005algorithmic} and \citet{park2022pac} for methodological details. For the multi-task variant, we set both hyperparameters to $0.3$ and artificially randomly split calibration tasks into subtasks of size $1000$ to match the experimental setup from \citet{park2022pac}.

The results in \cref{fig:CP,tab:results} reveal substantial differences between approaches. Plain probes achieve a mean false positive rate of $0.34$, exceeding the target threshold of $0.3$, with high variability ($80$th percentile: $0.69$). Standard conformal prediction reduces the mean false positive rate to $0.25$, approaching but not consistently achieving the target, with the $80$th percentile still reaching $0.47$. The multi-task variant achieves the strongest false positive rate control, with a mean of $0.09$ and $80$th percentile of $0.23$, successfully staying below the target threshold.
However, these improvements in false positive rate control come at substantial cost to recall. While plain probes achieve $0.52$ mean recall and standard conformal prediction reaches $0.56$, the multi-task variant falls to just $0.24$. This dramatic reduction means that this conservative approach correctly identifies only about one-fourth of true positives, illustrating the fundamental trade-off when truthfulness representations fail to generalize.
Moreover, this poor performance also reflects a deeper issue: conformal prediction assumes the scoring function $f(h)$ reliably ranks correctness across domains. However, our findings show that truthful and untruthful generations are only linearly separable within task-specific subspaces that vary significantly across tasks. When probes encounter out-of-distribution tasks, they become misaligned with actual correctness labels. To maintain the required false positive rate guarantees, conformal prediction must set extremely conservative thresholds, filtering out many correct answers not because they are ambiguous, but because the probe's direction no longer matches the task's geometry of truth. Our insight is that conformal prediction becomes overly conservative precisely because the underlying geometry fails to generalize.

\paragraph{Conclusion.}
The premise of the ``geometry of truth'' hypothesis is that one might be able to detect whether an LLM provides a correct answer when prompted with a question. These works claim that a simple classifier, taken as inputs the activations generated by the LLM as it produces its answer, can suffice to predict the correctness of the final answer. Although extremely appealing, we show in this work that such a promise may not be yet reliable enough as it fails to transfer across domains and tasks, notably if the domain on which the probe was trained is markedly different from that where the performance of the classifier is evaluated. We explain this failure to generalize by noticing that probes trained independently on various tasks have both low similarity and small feature overlap when trained with sparse regularizers. We have explored more advanced classification paradigms, such as mixture-of-probes, which could have been able to handle this heterogeneity, but we were not able to achieve reliable generalization. We conclude that LLMs likely have multiple geometries of truth, but that they are irreconcilable and highly task-dependent.

\bibliography{references}
\bibliographystyle{icml2025}

\newpage
\appendix

\section{Qwen-2.5 7B Instruct}
\subsection{Layer 28}

\begin{figure}[h]
    \centering
    \begin{subfigure}[t]{\linewidth}
    \includegraphics[width=\linewidth,trim={0.cm 0.cm 0cm 0.2cm},clip]{Figures/plots/Qwen2.5-7B-Instruct-28/l2-regularization/pairwise_stop_token_of_the_output_emb_28_auroc.pdf}
    \end{subfigure}
    \begin{subfigure}[t]{\linewidth}
    \includegraphics[width=\linewidth,trim={0.cm 0.cm 0.cm 0.2cm},clip]{Figures/plots/Qwen2.5-7B-Instruct-28/l2-regularization/pairwise_difference_stop_token_of_the_output_emb_28_auroc.pdf}
    \end{subfigure}
    \caption{AUROC of probes trained on different tasks on the stop token of the output on last layer. Rows correspond to evaluation tasks while columns correspond to training tasks. The second plot represents the difference between the probe trained on this task and probes trained on the other datasets. Results are averaged over 5 runs.}
\end{figure}

\begin{figure}[h]
    \centering
    \begin{subfigure}[t]{\linewidth}
    \includegraphics[width=\linewidth,trim={0.cm 0.cm 0cm 0.2cm},clip]{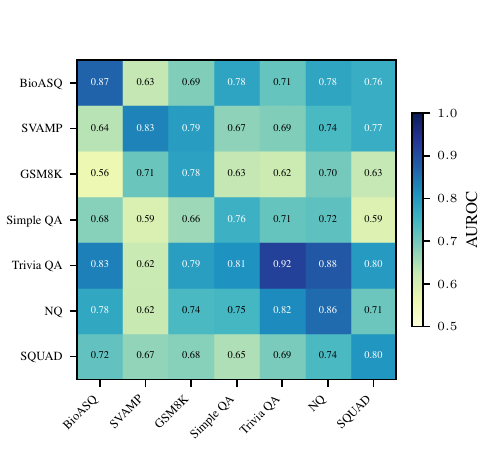}
    \end{subfigure}
    \begin{subfigure}[t]{\linewidth}
    \includegraphics[width=\linewidth,trim={0.cm 0.cm 0.cm 0.2cm},clip]{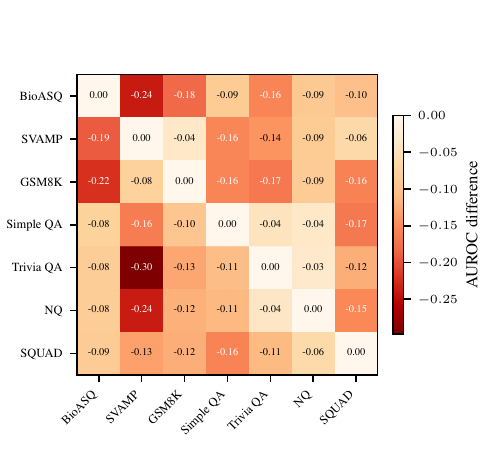}
    \end{subfigure}
    \caption{AUROC of probes trained on different tasks on the token before the stop token of the output on last layer. Rows correspond to evaluation tasks while columns correspond to training tasks. The second plot represents the difference between the probe trained on this task and probes trained on the other datasets. Results are averaged over 5 runs.}
\end{figure}

\begin{figure}[h]
    \centering
    \includegraphics[width=\columnwidth]{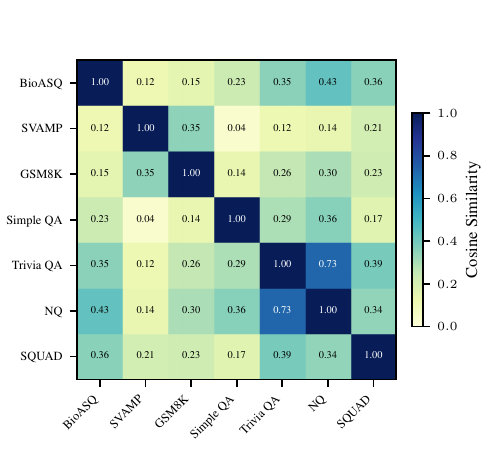}
    \caption{Cosine similarity between probes trained on different datasets using L2 regularization. Results are averaged over 5 runs.
    }
\end{figure}

\begin{figure}[h]
    \centering
    \includegraphics[width=\columnwidth]{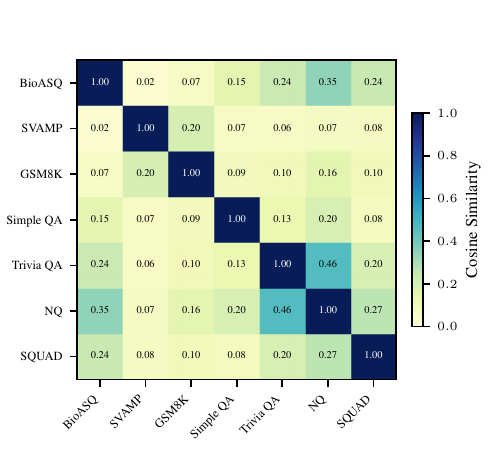}
    \caption{Cosine similarity between probes trained on different datasets using L2 regularization. Results are averaged over 5 runs.
    }
\end{figure}

\begin{figure}[h]
    \centering
    \includegraphics[width=\columnwidth]{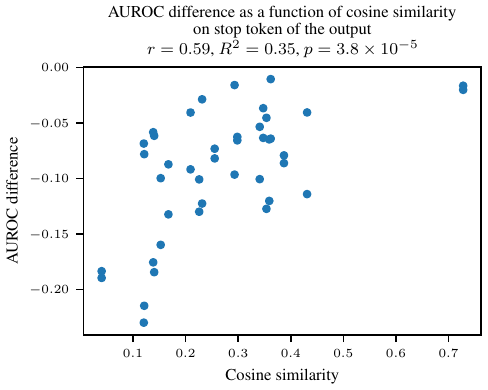}
    \caption{AUROC difference to probe trained on the right dataset as a function of cosine similarity between probes. Results are averaged over 5 runs.}
\end{figure}

\begin{figure}[h]
    \centering
    \includegraphics[width=\columnwidth]{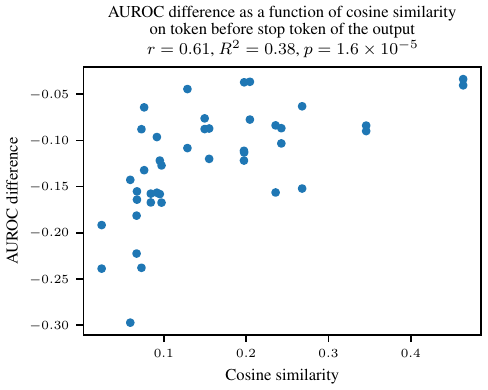}
    \caption{AUROC difference to probe trained on the right dataset as a function of cosine similarity between probes. Results are averaged over 5 runs.}
\end{figure}

\begin{figure}[h]
    \centering
    \includegraphics[width=\columnwidth]{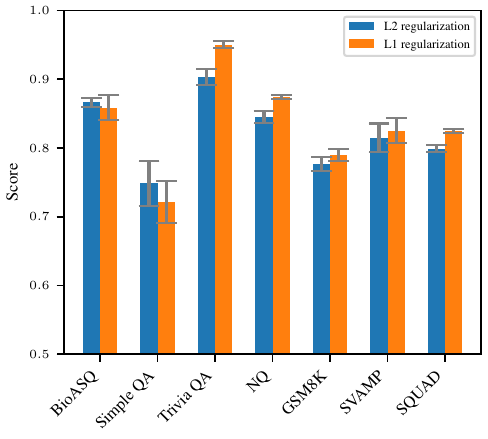}
    \caption{AUROC of linear probes at the stop token of the output, trained with either L1 or L2 regularization.}
\end{figure}

\begin{figure}[h]
    \centering
    \includegraphics[width=\columnwidth]{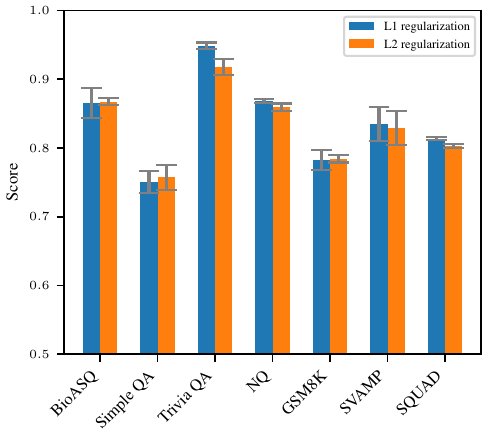}
    \caption{AUROC of linear probes at the token before the stop token of the output, trained with either L1 or L2 regularization.}
\end{figure}

\begin{figure}[h]
    \centering
    \includegraphics[width=\columnwidth]{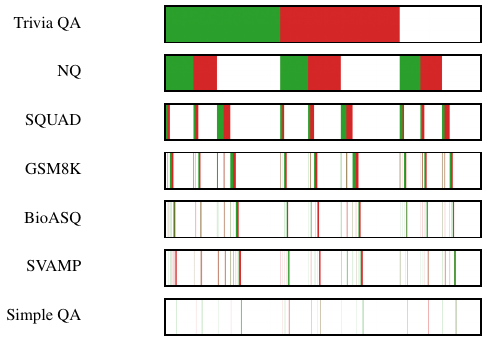}
    \caption{Signed support of sparse probes trained on different datasets at the stop token of the output, using L1 regularization at layer 28. Each row represents one probe trained on the corresponding dataset (y-axis labels). The x-axis shows the 3584 dimensions of the hidden state vector. Green indicates positive coefficients, red indicates negative coefficients, and white indicates zero coefficients (sparsity). Dimensions are sorted by sparsity level across all datasets, with the least sparse dimensions on the left and the most sparse on the right.}
\end{figure}

\begin{figure}[h]
    \centering
    \includegraphics[width=\columnwidth]{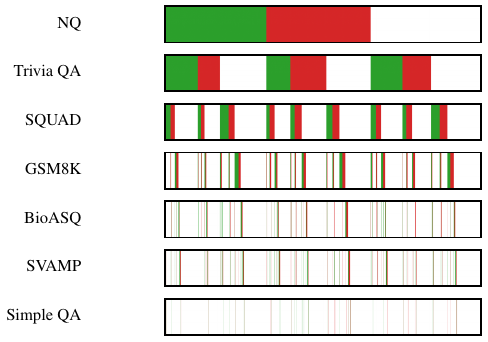}
    \caption{Signed support of sparse probes trained on different datasets at the token before the stop token of the output, using L1 regularization at layer 28. Each row represents one probe trained on the corresponding dataset (y-axis labels). The x-axis shows the 3584 dimensions of the hidden state vector. Green indicates positive coefficients, red indicates negative coefficients, and white indicates zero coefficients (sparsity). Dimensions are sorted by sparsity level across all datasets, with the least sparse dimensions on the left and the most sparse on the right.}
\end{figure}

\begin{figure}[h]
    \centering
    \includegraphics[width=\columnwidth]{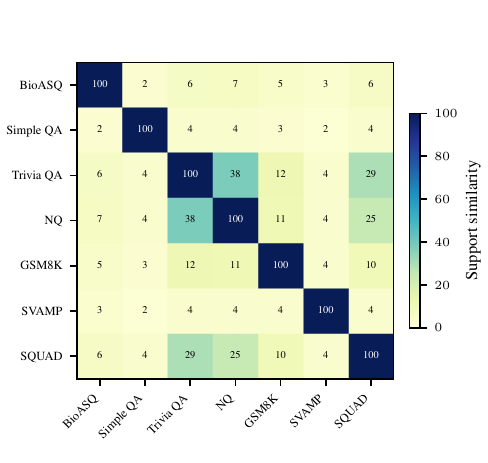}
    \caption{Support overlap between sparse probes trained on different datasets. Darker colors indicate higher overlap percentages. Task pairs with $>30\%$ overlap (TriviaQA, NQ, SimpleQA) correspond to successful cross-task generalization, while most pairs show $<15\%$ overlap, explaining generalization failure.}
\end{figure}

\begin{figure}[h]
    \centering
    \includegraphics[width=\columnwidth]{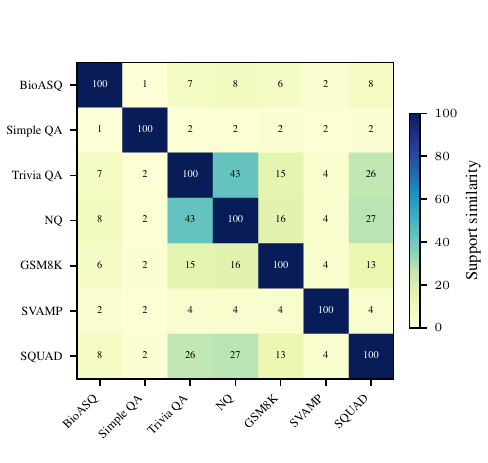}
    \caption{Support overlap between sparse probes trained on different datasets. Darker colors indicate higher overlap percentages. Task pairs with $>30\%$ overlap (TriviaQA, NQ, SimpleQA) correspond to successful cross-task generalization, while most pairs show $<15\%$ overlap, explaining generalization failure.}
\end{figure}

\begin{figure}[h]
    \centering
    \includegraphics[width=\columnwidth]{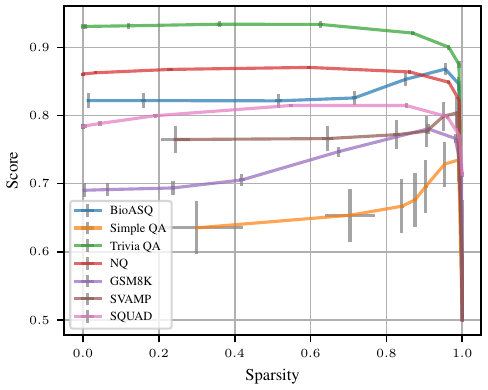}
    \caption{AUROC of probes trained using L1 regularisation as a function of the sparsity level on the stop token of the output.}
\end{figure}

\begin{figure}[h]
    \centering
    \includegraphics[width=\columnwidth]{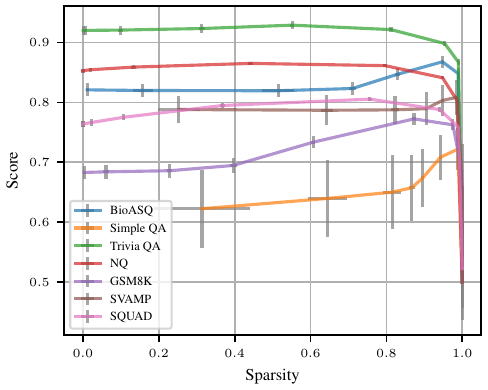}
    \caption{AUROC of probes trained using L1 regularisation as a function of the sparsity level on the token before the stop token of the output.}
\end{figure}

\begin{figure}[h]
    \centering
    \begin{subfigure}[t]{\columnwidth}
    \includegraphics[width=\textwidth]{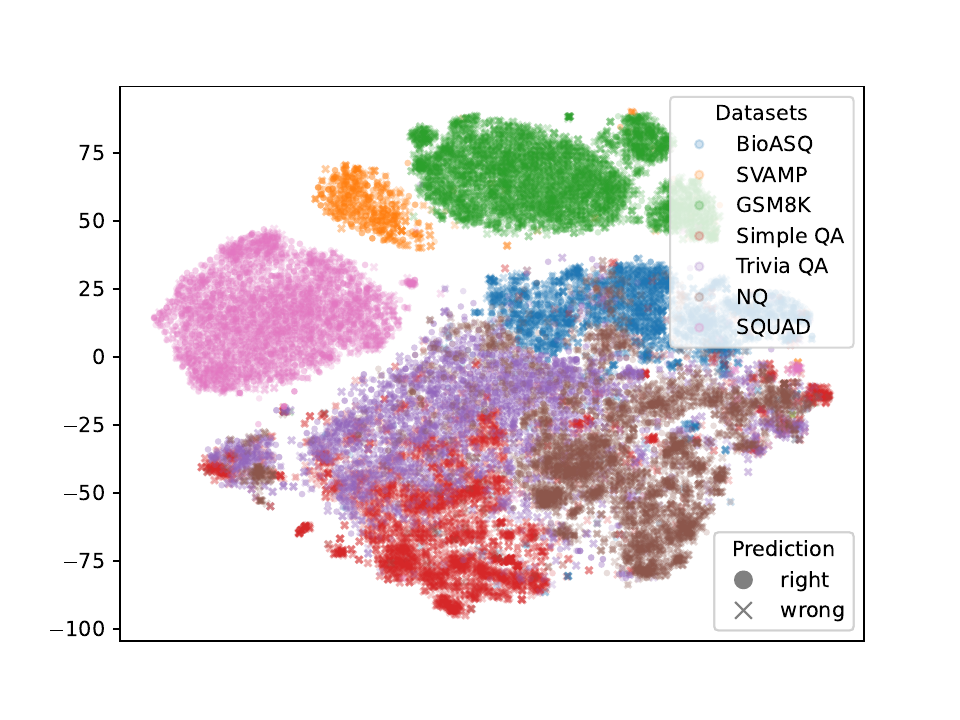}
    \end{subfigure}
    \caption{t-SNE plots of the hidden space at layer 28 at the stop token of the output.}
\end{figure}

\begin{figure}[h]
    \centering
    \begin{subfigure}[t]{\columnwidth}
    \includegraphics[width=\textwidth]{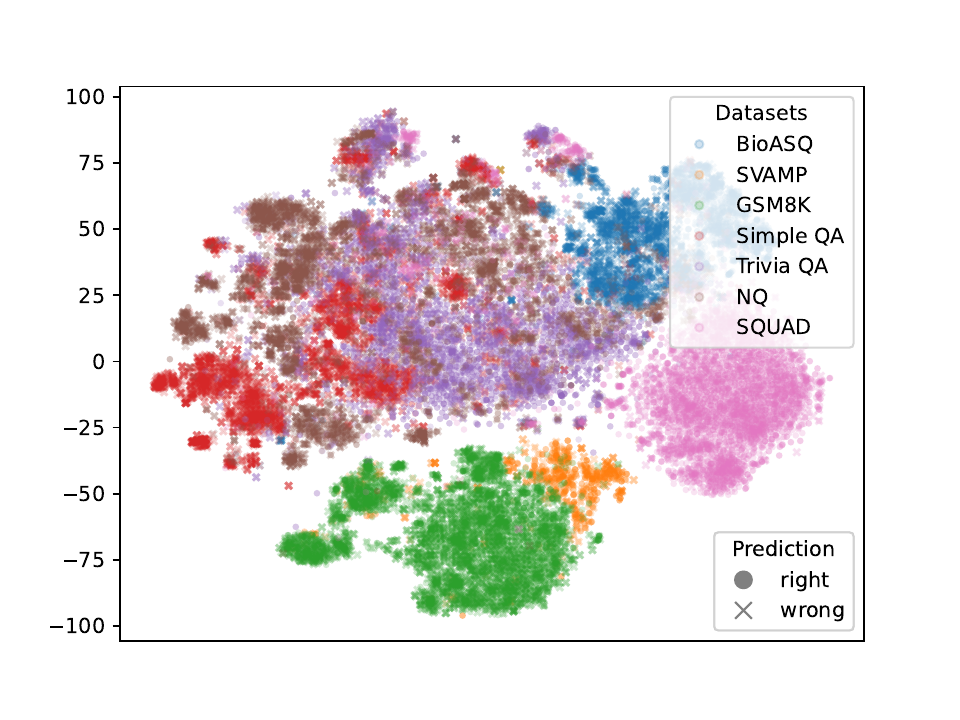}
    \end{subfigure}
    \caption{t-SNE plots of the hidden space at layer 28 at the token before stop token of the output.}
\end{figure}

\begin{figure*}[!h]
    \centering
    \includegraphics[width=\linewidth]{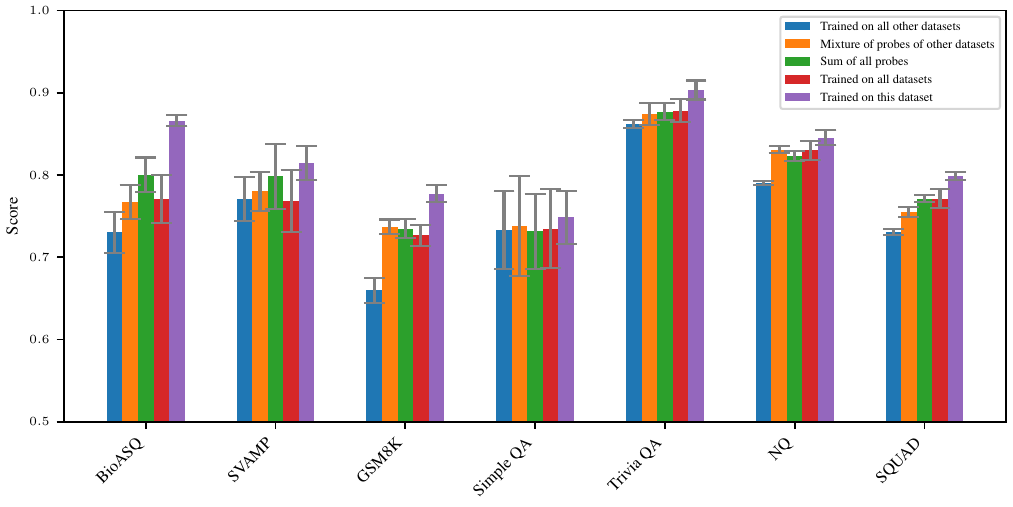}
    \caption{AUROC of linear probes at the stop token of the output in the multi-task setting, using L2 regularization.}
\end{figure*}

\begin{figure*}[!h]
    \centering
    \includegraphics[width=\linewidth]{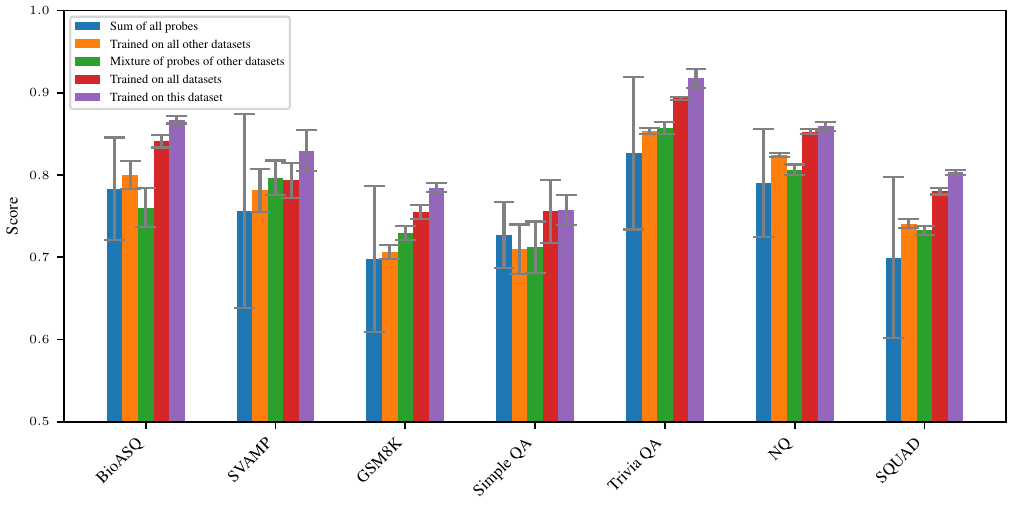}
    \caption{AUROC of linear probes at the token before the stop token of the output in the multi-task setting, using L2 regularization.}
\end{figure*}

\FloatBarrier

\subsection{Layer 21}

\begin{figure}[h]
    \centering
    \begin{subfigure}[t]{\linewidth}
    \includegraphics[width=\linewidth,trim={0.cm 0.cm 0cm 0.2cm},clip]{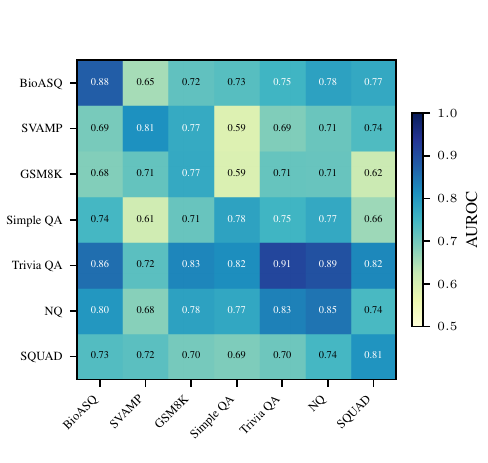}
    \end{subfigure}
    \begin{subfigure}[t]{\linewidth}
    \includegraphics[width=\linewidth,trim={0.cm 0.cm 0.cm 0.2cm},clip]{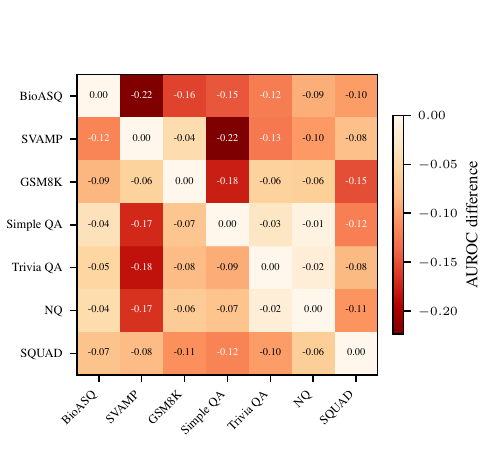}
    \end{subfigure}
    \caption{AUROC of probes trained on different tasks on the stop token of the output on last layer. Rows correspond to evaluation tasks while columns correspond to training tasks. The second plot represents the difference between the probe trained on this task and probes trained on the other datasets. Results are averaged over 5 runs.}
\end{figure}

\begin{figure}[h]
    \centering
    \begin{subfigure}[t]{\linewidth}
    \includegraphics[width=\linewidth,trim={0.cm 0.cm 0cm 0.2cm},clip]{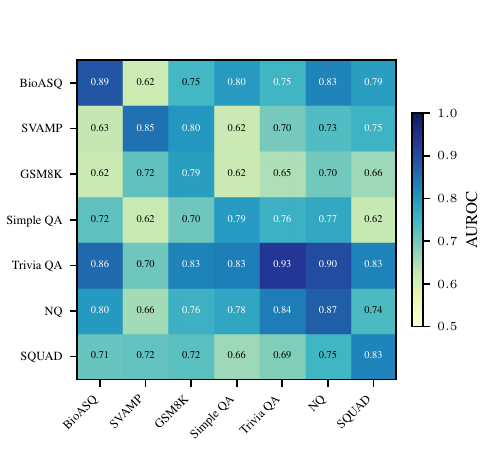}
    \end{subfigure}
    \begin{subfigure}[t]{\linewidth}
    \includegraphics[width=\linewidth,trim={0.cm 0.cm 0.cm 0.2cm},clip]{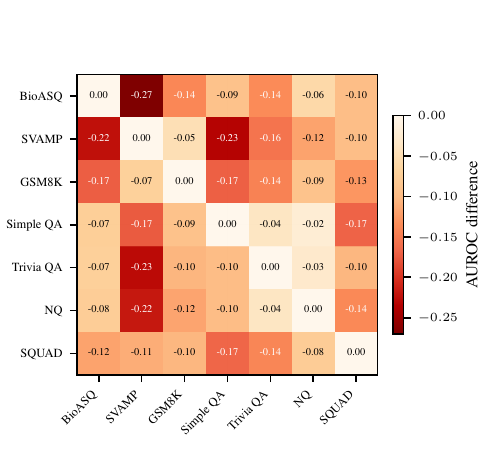}
    \end{subfigure}
    \caption{AUROC of probes trained on different tasks on the token before the stop token of the output on last layer. Rows correspond to evaluation tasks while columns correspond to training tasks. The second plot represents the difference between the probe trained on this task and probes trained on the other datasets. Results are averaged over 5 runs.}
\end{figure}

\begin{figure}[h]
    \centering
    \includegraphics[width=\columnwidth]{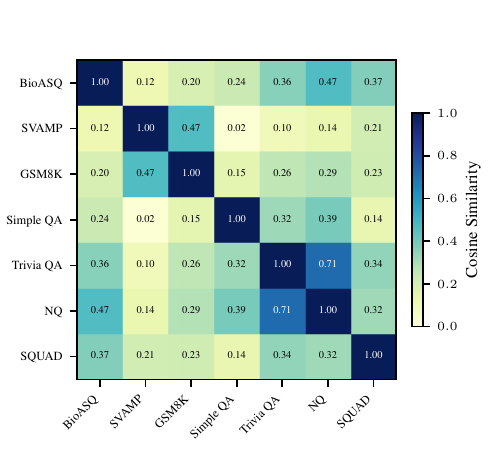}
    \caption{Cosine similarity between probes trained on different datasets using L2 regularization. Results are averaged over 5 runs.
    }
\end{figure}

\begin{figure}[h]
    \centering
    \includegraphics[width=\columnwidth]{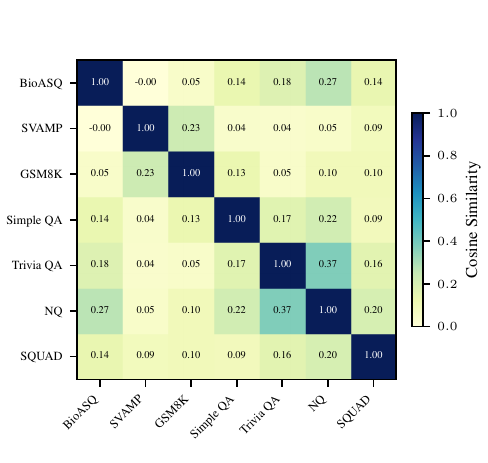}
    \caption{Cosine similarity between probes trained on different datasets using L2 regularization. Results are averaged over 5 runs.
    }
\end{figure}

\begin{figure}[h]
    \centering
    \includegraphics[width=\columnwidth]{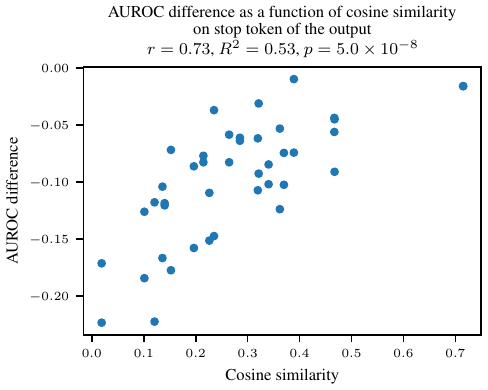}
    \caption{AUROC difference to probe trained on the right dataset as a function of cosine similarity between probes. Results are averaged over 5 runs.}
\end{figure}

\begin{figure}[h]
    \centering
    \includegraphics[width=\columnwidth]{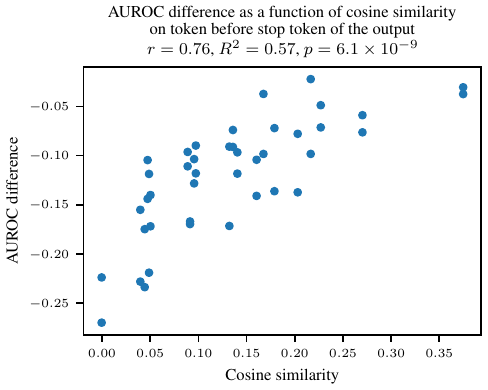}
    \caption{AUROC difference to probe trained on the right dataset as a function of cosine similarity between probes. Results are averaged over 5 runs.}
\end{figure}

\begin{figure}[h]
    \centering
    \includegraphics[width=\columnwidth]{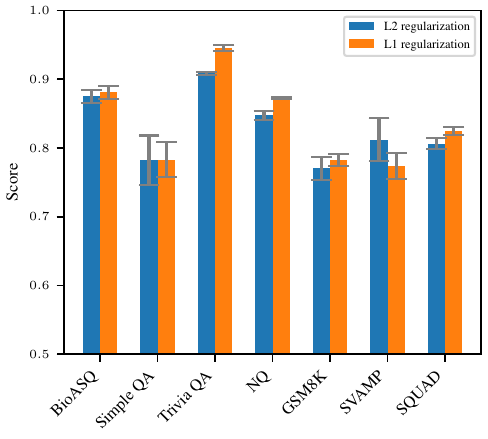}
    \caption{AUROC of linear probes at the stop token of the output, trained with either L1 or L2 regularization.}
\end{figure}

\begin{figure}[h]
    \centering
    \includegraphics[width=\columnwidth]{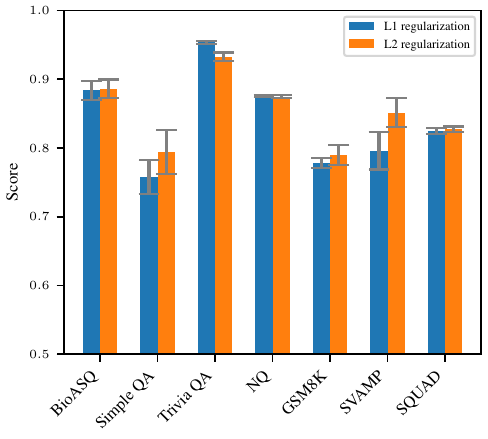}
    \caption{AUROC of linear probes at the token before the stop token of the output, trained with either L1 or L2 regularization.}
\end{figure}

\begin{figure}[h]
    \centering
    \includegraphics[width=\columnwidth]{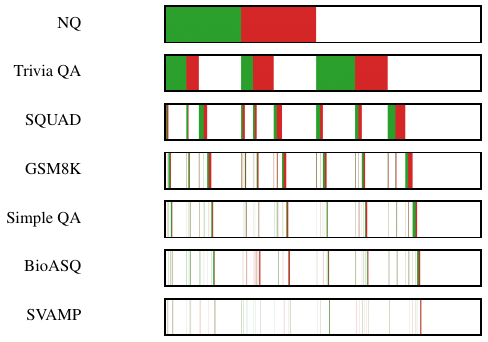}
    \caption{Signed support of sparse probes trained on different datasets at the stop token of the output, using L1 regularization at layer 21. Each row represents one probe trained on the corresponding dataset (y-axis labels). The x-axis shows the 3584 dimensions of the hidden state vector. Green indicates positive coefficients, red indicates negative coefficients, and white indicates zero coefficients (sparsity). Dimensions are sorted by sparsity level across all datasets, with the least sparse dimensions on the left and the most sparse on the right.}
\end{figure}

\begin{figure}[h]
    \centering
    \includegraphics[width=\columnwidth]{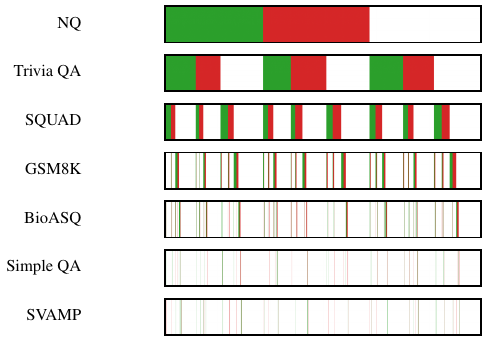}
    \caption{Signed support of sparse probes trained on different datasets at the token before the stop token of the output, using L1 regularization at layer 21. Each row represents one probe trained on the corresponding dataset (y-axis labels). The x-axis shows the 3584 dimensions of the hidden state vector. Green indicates positive coefficients, red indicates negative coefficients, and white indicates zero coefficients (sparsity). Dimensions are sorted by sparsity level across all datasets, with the least sparse dimensions on the left and the most sparse on the right.}
\end{figure}

\begin{figure}[h]
    \centering
    \includegraphics[width=\columnwidth]{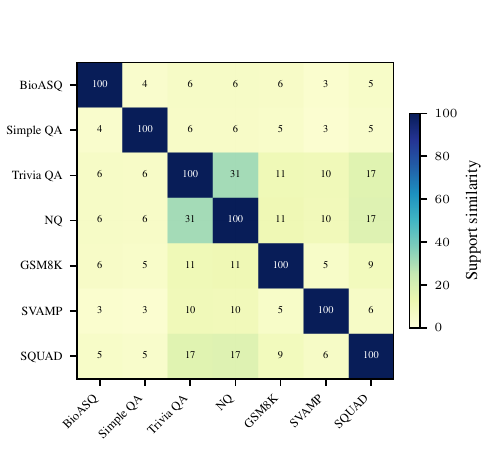}
    \caption{Support overlap between sparse probes trained on different datasets. Darker colors indicate higher overlap percentages. Task pairs with $>30\%$ overlap (TriviaQA, NQ, SimpleQA) correspond to successful cross-task generalization, while most pairs show $<15\%$ overlap, explaining generalization failure.}
\end{figure}

\begin{figure}[h]
    \centering
    \includegraphics[width=\columnwidth]{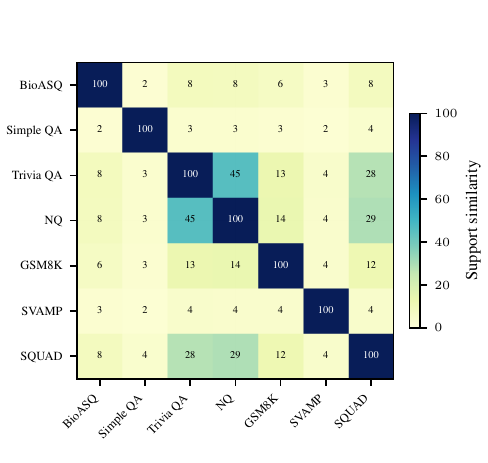}
    \caption{Support overlap between sparse probes trained on different datasets. Darker colors indicate higher overlap percentages. Task pairs with $>30\%$ overlap (TriviaQA, NQ, SimpleQA) correspond to successful cross-task generalization, while most pairs show $<15\%$ overlap, explaining generalization failure.}
\end{figure}

\begin{figure}[h]
    \centering
    \includegraphics[width=\columnwidth]{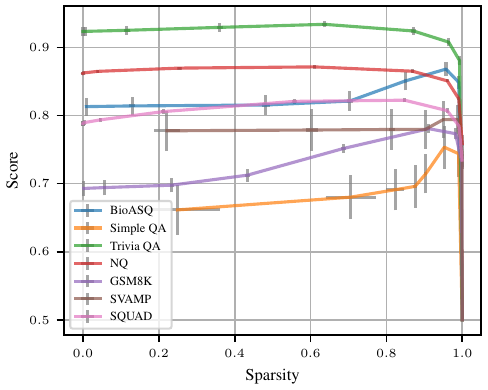}
    \caption{AUROC of probes trained using L1 regularisation as a function of the sparsity level on the stop token of the output.}
\end{figure}

\begin{figure}[h]
    \centering
    \includegraphics[width=\columnwidth]{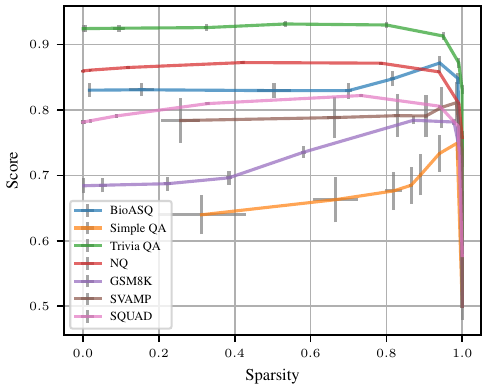}
    \caption{AUROC of probes trained using L1 regularisation as a function of the sparsity level on the token before the stop token of the output.}
\end{figure}

\begin{figure}[h]
    \centering
    \begin{subfigure}[t]{\columnwidth}
    \includegraphics[width=\textwidth]{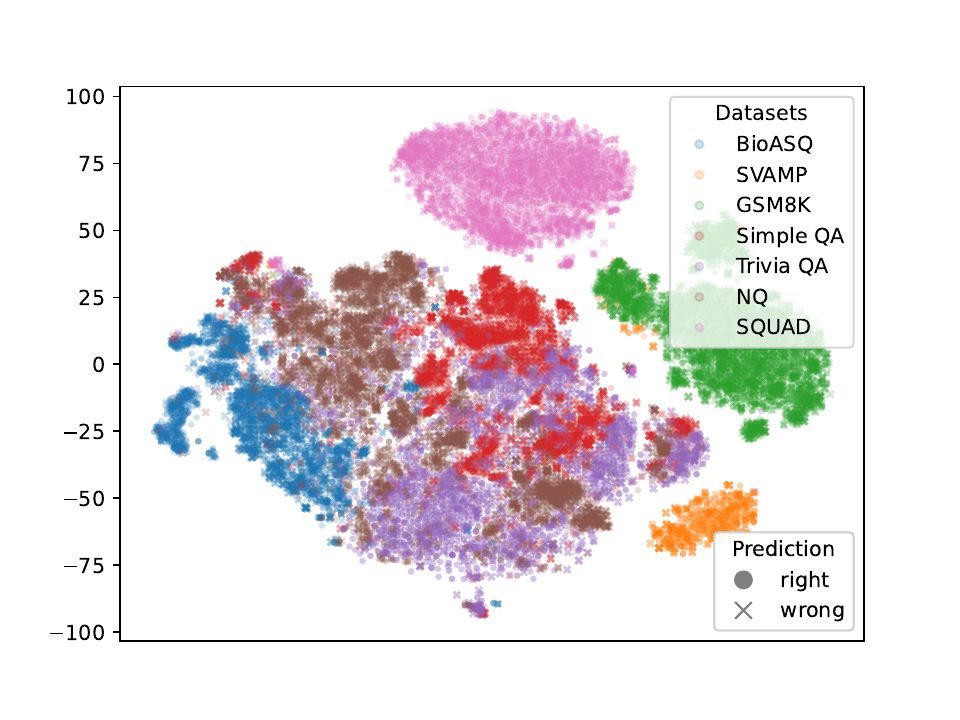}
    \end{subfigure}
    \caption{t-SNE plots of the hidden space at layer 21 at the stop token of the output.}
\end{figure}

\begin{figure}[h]
    \centering
    \begin{subfigure}[t]{\columnwidth}
    \includegraphics[width=\textwidth]{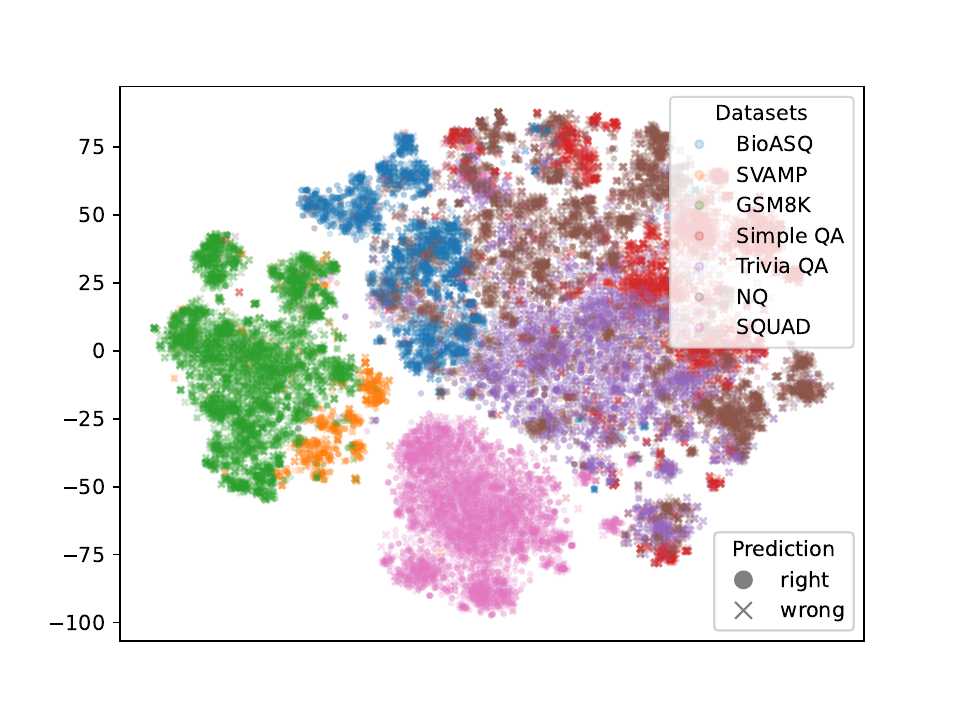}
    \end{subfigure}
    \caption{t-SNE plots of the hidden space at layer 21 at the token before stop token of the output.}
\end{figure}

\begin{figure*}[!h]
    \centering
    \includegraphics[width=\linewidth]{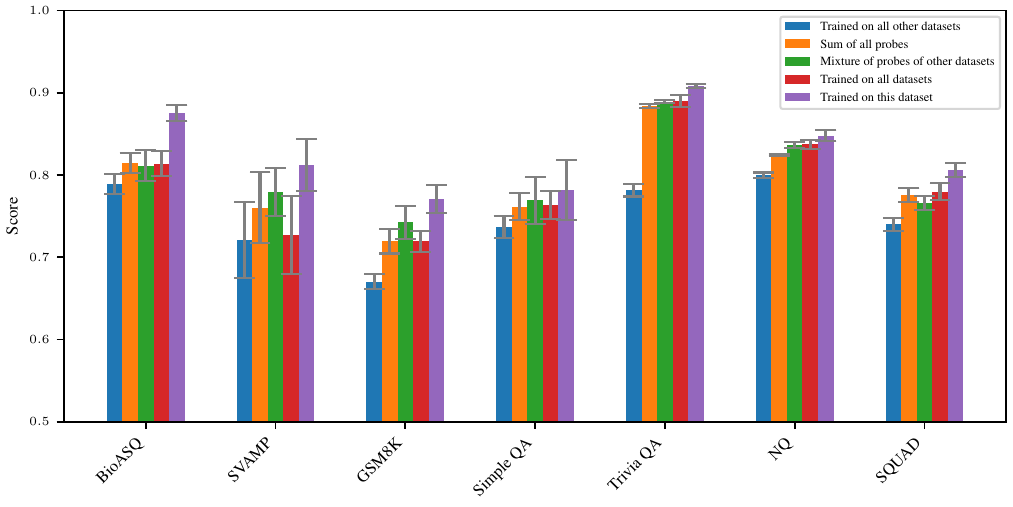}
    \caption{AUROC of linear probes at the stop token of the output in the multi-task setting, using L2 regularization.}
\end{figure*}

\begin{figure*}[!h]
    \centering
    \includegraphics[width=\linewidth]{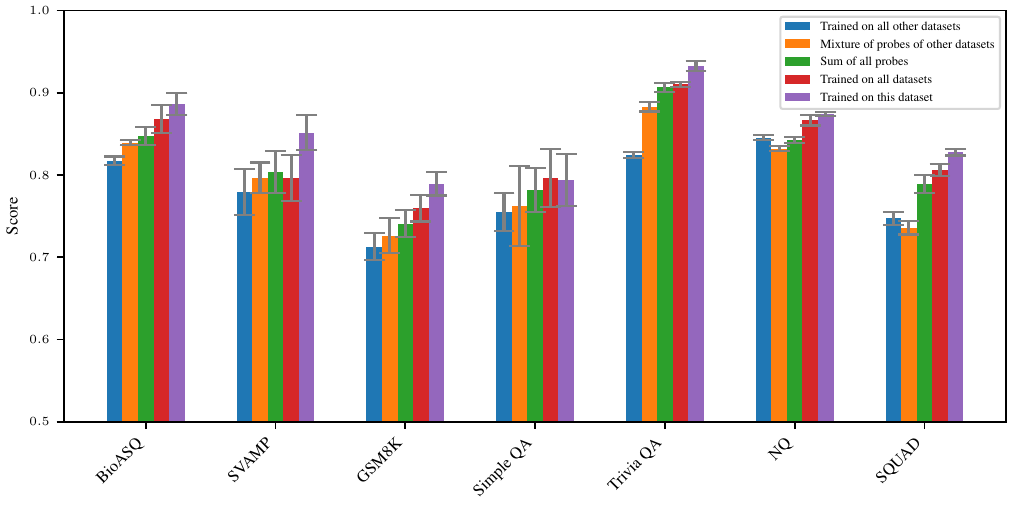}
    \caption{AUROC of linear probes at the token before the stop token of the output in the multi-task setting, using L2 regularization.}
\end{figure*}

\FloatBarrier

\section{Llama 3.1 8B Instruct}
\subsection{Layer 32}

\begin{figure}[h]
    \centering
    \begin{subfigure}[t]{\linewidth}
    \includegraphics[width=\linewidth,trim={0.cm 0.cm 0cm 0.2cm},clip]{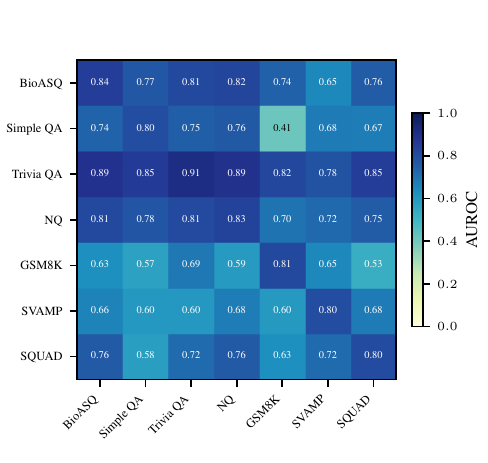}
    \end{subfigure}
    \begin{subfigure}[t]{\linewidth}
    \includegraphics[width=\linewidth,trim={0.cm 0.cm 0.cm 0.2cm},clip]{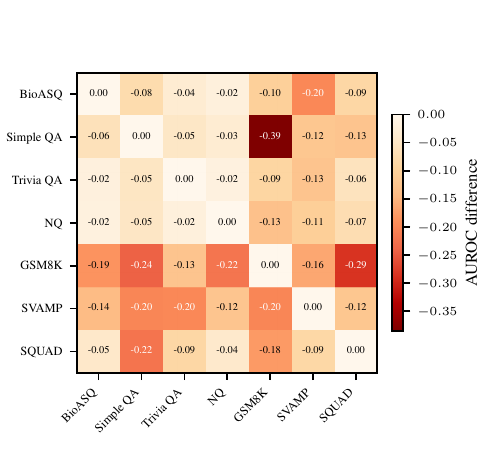}
    \end{subfigure}
    \caption{AUROC of probes trained on different tasks on the stop token of the output on last layer. Rows correspond to evaluation tasks while columns correspond to training tasks. The second plot represents the difference between the probe trained on this task and probes trained on the other datasets. Results are averaged over 5 runs.}
\end{figure}

\begin{figure}[h]
    \centering
    \begin{subfigure}[t]{\linewidth}
    \includegraphics[width=\linewidth,trim={0.cm 0.cm 0cm 0.2cm},clip]{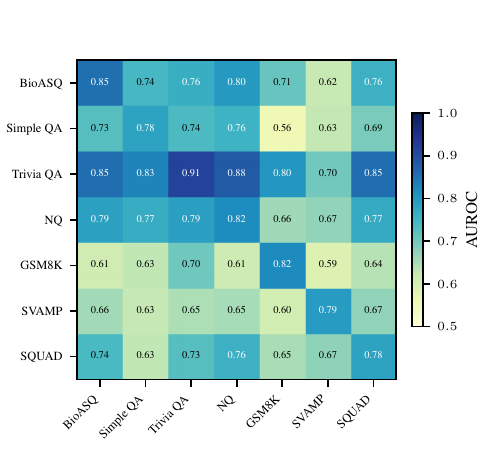}
    \end{subfigure}
    \begin{subfigure}[t]{\linewidth}
    \includegraphics[width=\linewidth,trim={0.cm 0.cm 0.cm 0.2cm},clip]{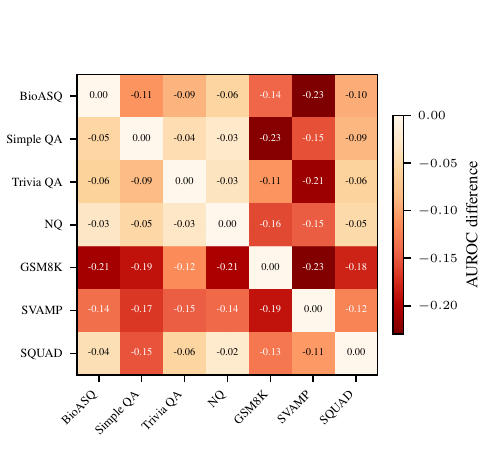}
    \end{subfigure}
    \caption{AUROC of probes trained on different tasks on the token before the stop token of the output on last layer. Rows correspond to evaluation tasks while columns correspond to training tasks. The second plot represents the difference between the probe trained on this task and probes trained on the other datasets. Results are averaged over 5 runs.}
\end{figure}

\begin{figure}[h]
    \centering
    \includegraphics[width=\columnwidth]{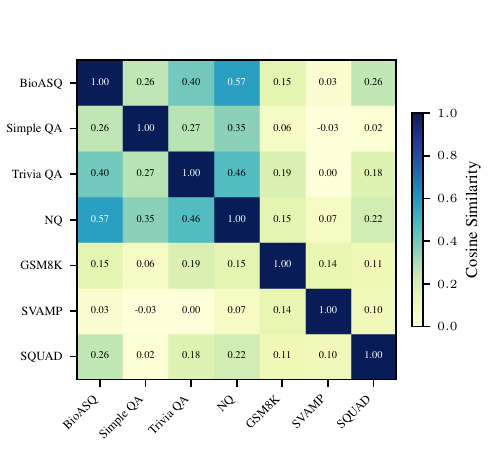}
    \caption{Cosine similarity between probes trained on different datasets using L2 regularization. Results are averaged over 5 runs.
    }
\end{figure}

\begin{figure}[h]
    \centering
    \includegraphics[width=\columnwidth]{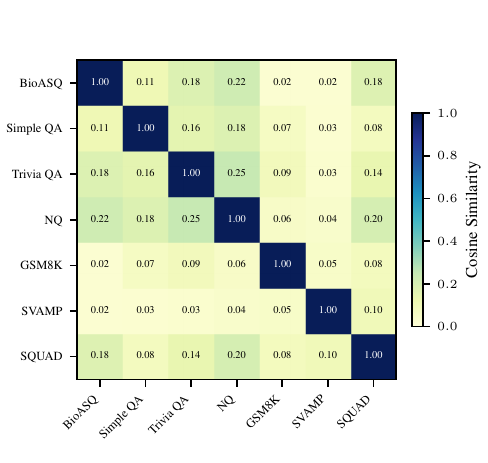}
    \caption{Cosine similarity between probes trained on different datasets using L2 regularization. Results are averaged over 5 runs.
    }
\end{figure}

\begin{figure}[h]
    \centering
    \includegraphics[width=\columnwidth]{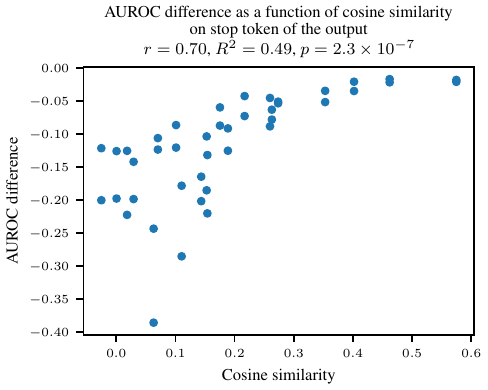}
    \caption{AUROC difference to probe trained on the right dataset as a function of cosine similarity between probes. Results are averaged over 5 runs.}
\end{figure}

\begin{figure}[h]
    \centering
    \includegraphics[width=\columnwidth]{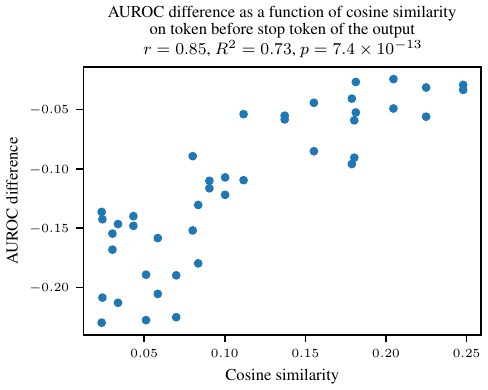}
    \caption{AUROC difference to probe trained on the right dataset as a function of cosine similarity between probes. Results are averaged over 5 runs.}
\end{figure}

\begin{figure}[h]
    \centering
    \includegraphics[width=\columnwidth]{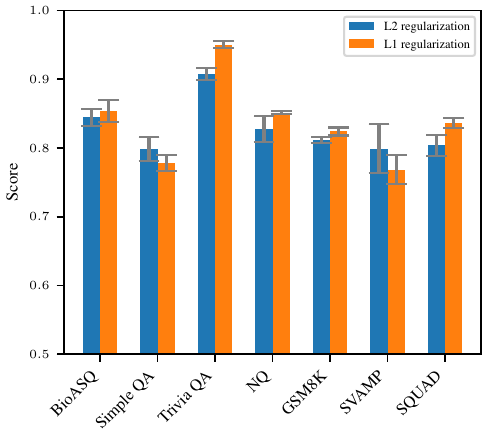}
    \caption{AUROC of linear probes at the stop token of the output, trained with either L1 or L2 regularization.}
\end{figure}

\begin{figure}[h]
    \centering
    \includegraphics[width=\columnwidth]{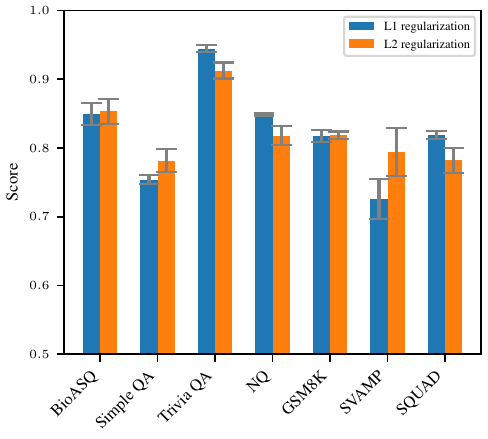}
    \caption{AUROC of linear probes at the token before the stop token of the output, trained with either L1 or L2 regularization.}
\end{figure}

\begin{figure}[h]
    \centering
    \includegraphics[width=\columnwidth]{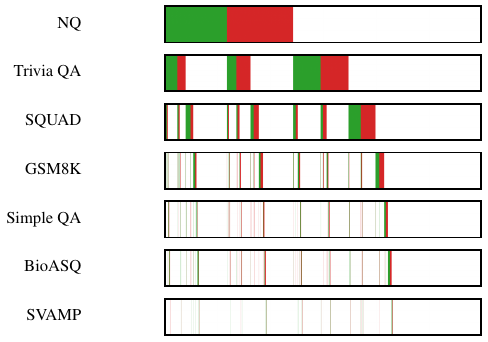}
    \caption{Signed support of sparse probes trained on different datasets at the stop token of the output, using L1 regularization at layer 32. Each row represents one probe trained on the corresponding dataset (y-axis labels). The x-axis shows the 3584 dimensions of the hidden state vector. Green indicates positive coefficients, red indicates negative coefficients, and white indicates zero coefficients (sparsity). Dimensions are sorted by sparsity level across all datasets, with the least sparse dimensions on the left and the most sparse on the right.}
\end{figure}

\begin{figure}[h]
    \centering
    \includegraphics[width=\columnwidth]{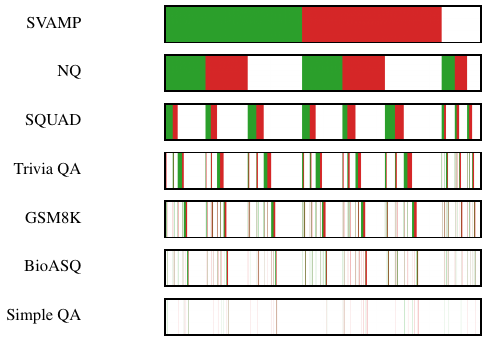}
    \caption{Signed support of sparse probes trained on different datasets at the token before the stop token of the output, using L1 regularization at layer 32. Each row represents one probe trained on the corresponding dataset (y-axis labels). The x-axis shows the 3584 dimensions of the hidden state vector. Green indicates positive coefficients, red indicates negative coefficients, and white indicates zero coefficients (sparsity). Dimensions are sorted by sparsity level across all datasets, with the least sparse dimensions on the left and the most sparse on the right.}
\end{figure}

\begin{figure}[h]
    \centering
    \includegraphics[width=\columnwidth]{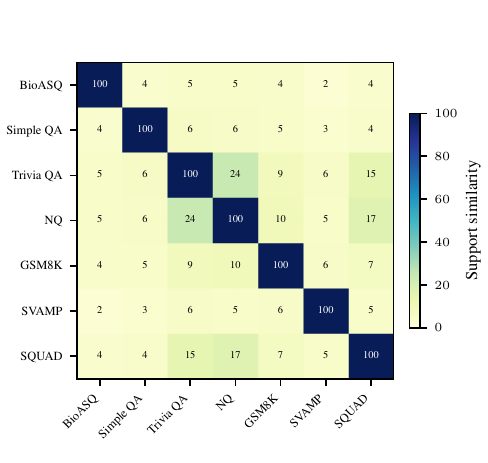}
    \caption{Support overlap between sparse probes trained on different datasets. Darker colors indicate higher overlap percentages. Task pairs with $>30\%$ overlap (TriviaQA, NQ, SimpleQA) correspond to successful cross-task generalization, while most pairs show $<15\%$ overlap, explaining generalization failure.}
\end{figure}

\begin{figure}[h]
    \centering
    \includegraphics[width=\columnwidth]{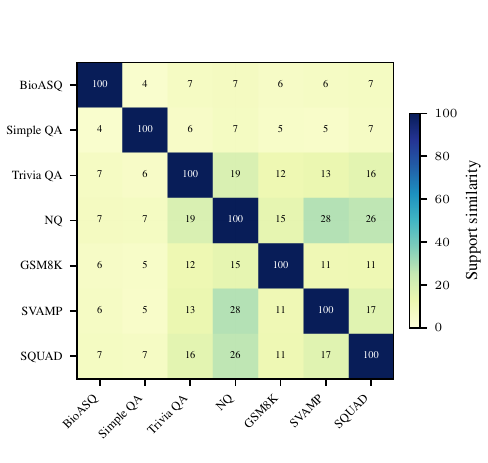}
    \caption{Support overlap between sparse probes trained on different datasets. Darker colors indicate higher overlap percentages. Task pairs with $>30\%$ overlap (TriviaQA, NQ, SimpleQA) correspond to successful cross-task generalization, while most pairs show $<15\%$ overlap, explaining generalization failure.}
\end{figure}

\begin{figure}[h]
    \centering
    \includegraphics[width=\columnwidth]{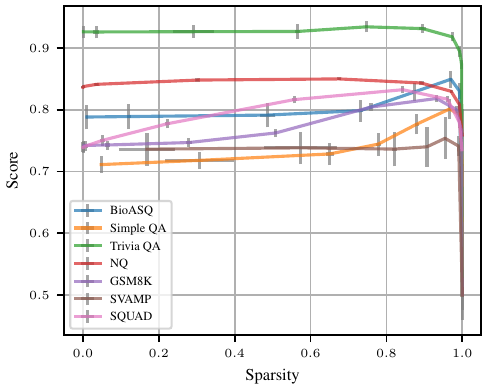}
    \caption{AUROC of probes trained using L1 regularisation as a function of the sparsity level on the stop token of the output.}
\end{figure}

\begin{figure}[h]
    \centering
    \includegraphics[width=\columnwidth]{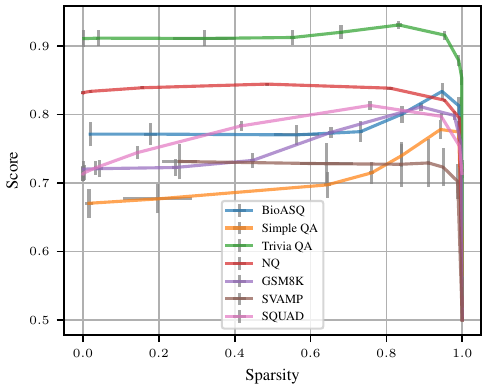}
    \caption{AUROC of probes trained using L1 regularisation as a function of the sparsity level on the token before the stop token of the output.}
\end{figure}

\begin{figure}[h]
    \centering
    \begin{subfigure}[t]{\columnwidth}
    \includegraphics[width=\textwidth]{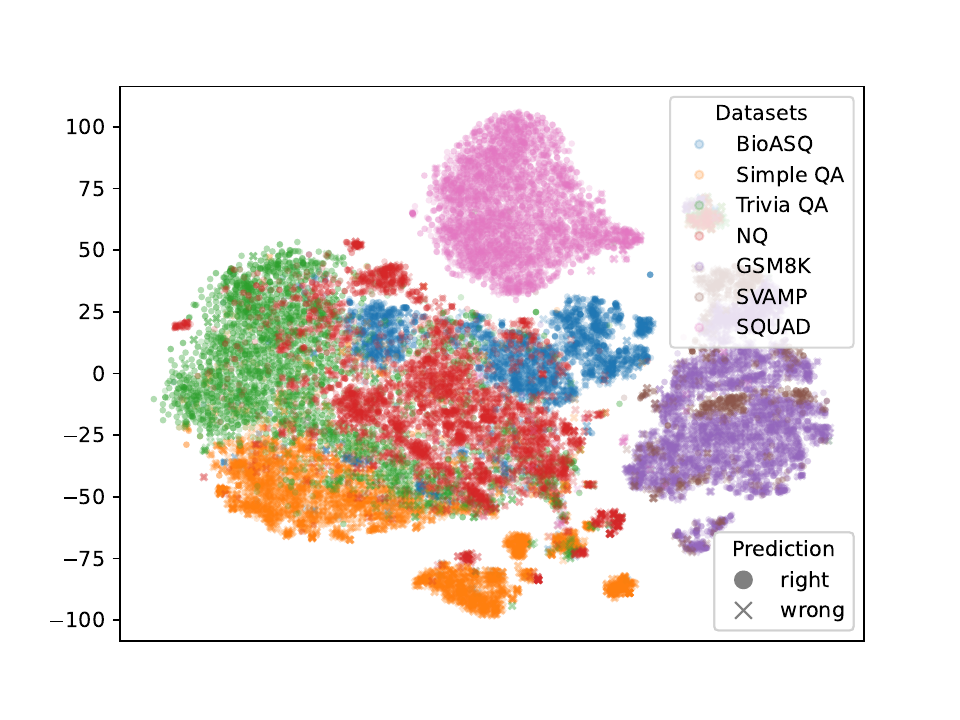}
    \end{subfigure}
    \caption{t-SNE plots of the hidden space at layer 32 at the stop token of the output.}
\end{figure}

\begin{figure}[h]
    \centering
    \begin{subfigure}[t]{\columnwidth}
    \includegraphics[width=\textwidth]{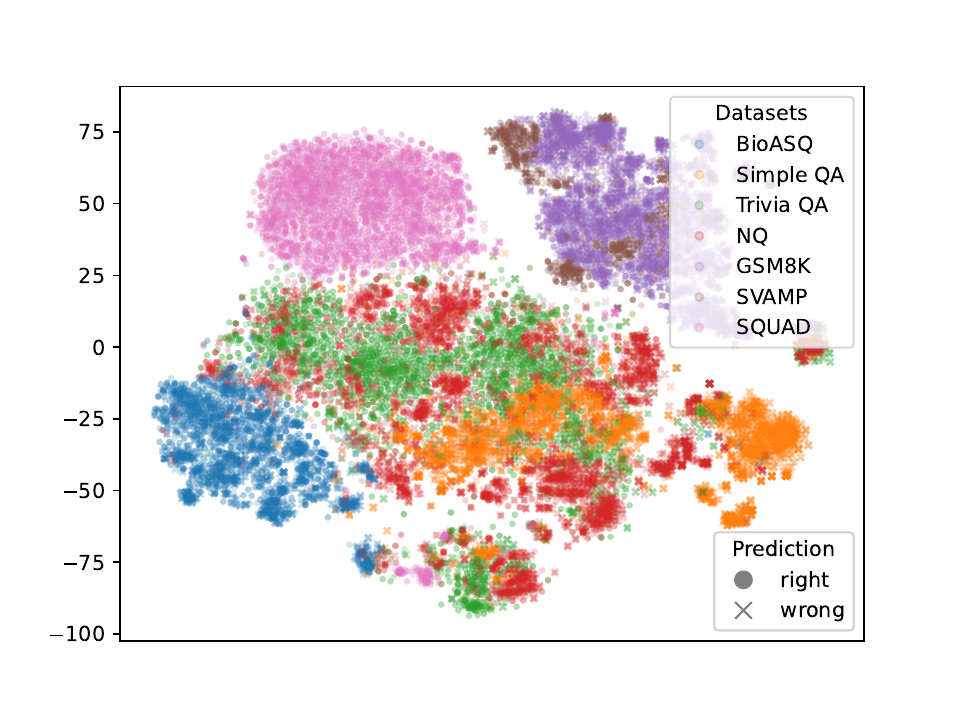}
    \end{subfigure}
    \caption{t-SNE plots of the hidden space at layer 32 at the token before stop token of the output.}
\end{figure}

\begin{figure*}[!h]
    \centering
    \includegraphics[width=\linewidth]{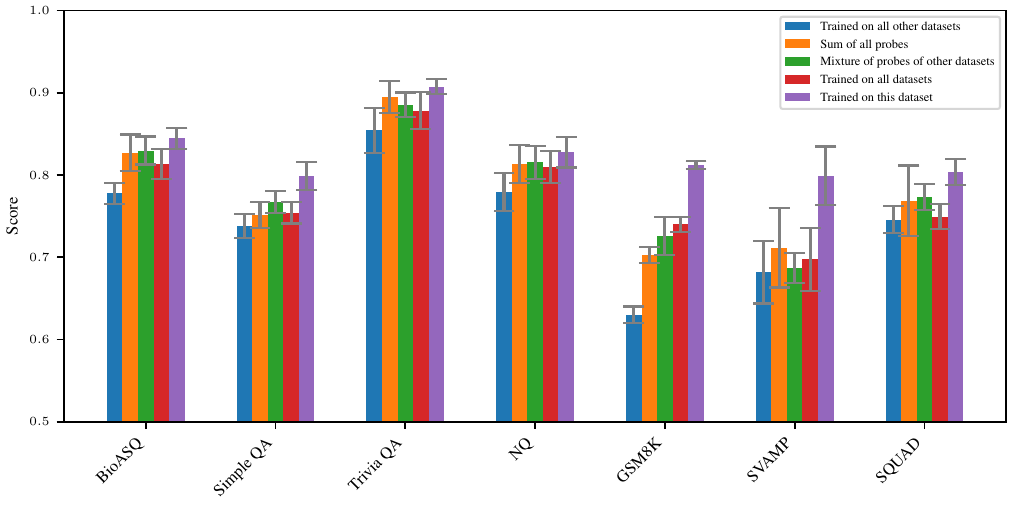}
    \caption{AUROC of linear probes at the stop token of the output in the multi-task setting, using L2 regularization.}
\end{figure*}

\begin{figure*}[!h]
    \centering
    \includegraphics[width=\linewidth]{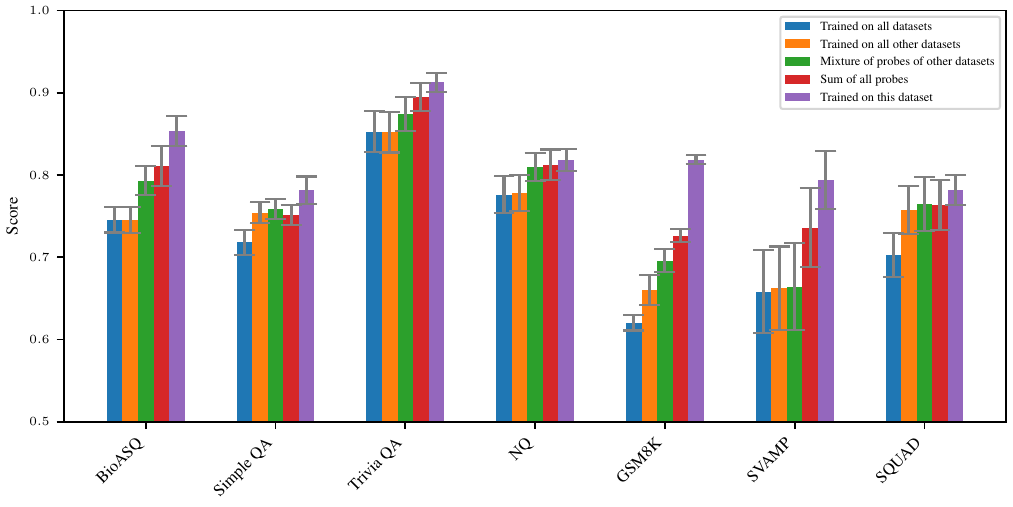}
    \caption{AUROC of linear probes at the token before the stop token of the output in the multi-task setting, using L2 regularization.}
\end{figure*}

\FloatBarrier

\subsection{Layer 24}

\begin{figure}[h]
    \centering
    \begin{subfigure}[t]{\linewidth}
    \includegraphics[width=\linewidth,trim={0.cm 0.cm 0cm 0.2cm},clip]{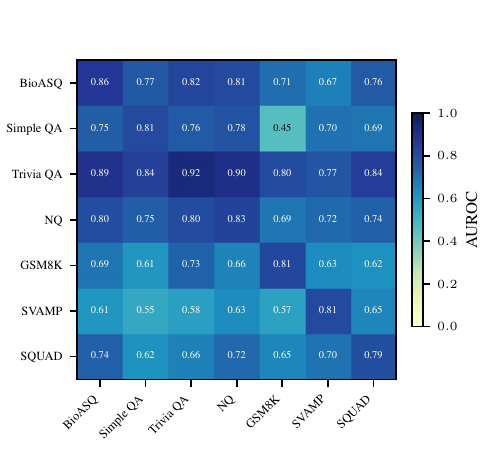}
    \end{subfigure}
    \begin{subfigure}[t]{\linewidth}
    \includegraphics[width=\linewidth,trim={0.cm 0.cm 0.cm 0.2cm},clip]{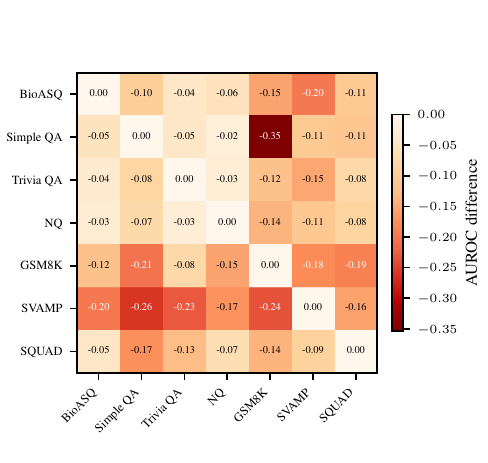}
    \end{subfigure}
    \caption{AUROC of probes trained on different tasks on the stop token of the output on last layer. Rows correspond to evaluation tasks while columns correspond to training tasks. The second plot represents the difference between the probe trained on this task and probes trained on the other datasets. Results are averaged over 5 runs.}
\end{figure}

\begin{figure}[h]
    \centering
    \begin{subfigure}[t]{\linewidth}
    \includegraphics[width=\linewidth,trim={0.cm 0.cm 0cm 0.2cm},clip]{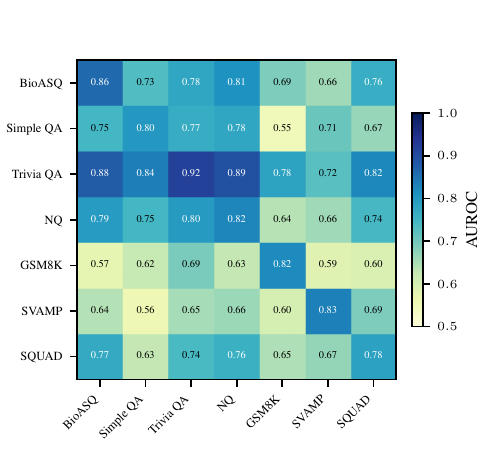}
    \end{subfigure}
    \begin{subfigure}[t]{\linewidth}
    \includegraphics[width=\linewidth,trim={0.cm 0.cm 0.cm 0.2cm},clip]{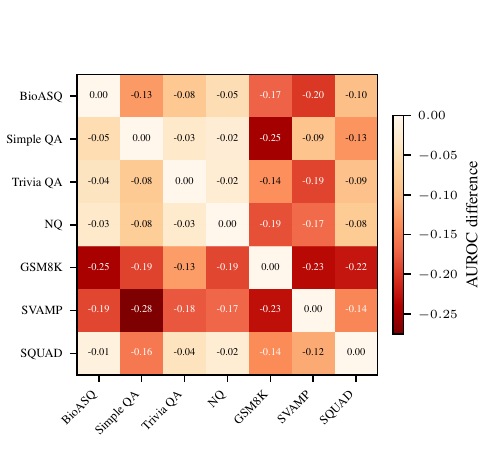}
    \end{subfigure}
    \caption{AUROC of probes trained on different tasks on the token before the stop token of the output on last layer. Rows correspond to evaluation tasks while columns correspond to training tasks. The second plot represents the difference between the probe trained on this task and probes trained on the other datasets. Results are averaged over 5 runs.}
\end{figure}

\begin{figure}[h]
    \centering
    \includegraphics[width=\columnwidth]{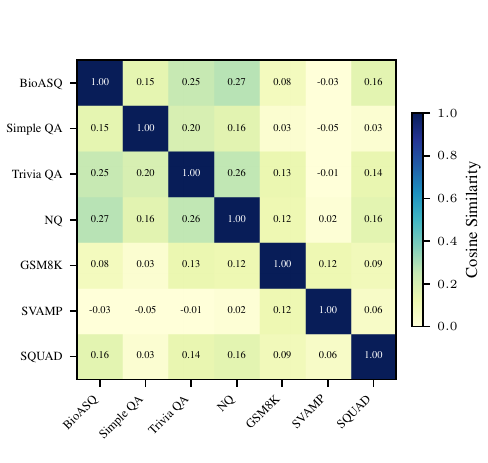}
    \caption{Cosine similarity between probes trained on different datasets using L2 regularization. Results are averaged over 5 runs.
    }
\end{figure}

\begin{figure}[h]
    \centering
    \includegraphics[width=\columnwidth]{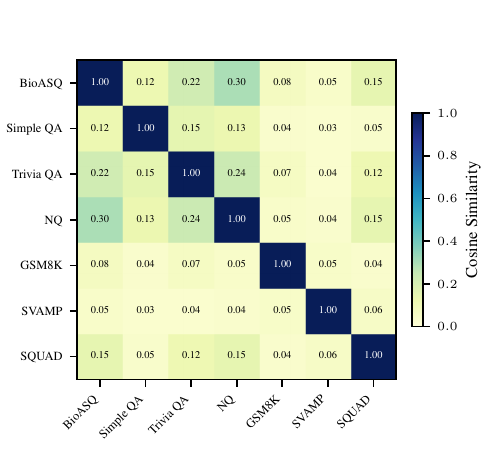}
    \caption{Cosine similarity between probes trained on different datasets using L2 regularization. Results are averaged over 5 runs.
    }
\end{figure}

\begin{figure}[h]
    \centering
    \includegraphics[width=\columnwidth]{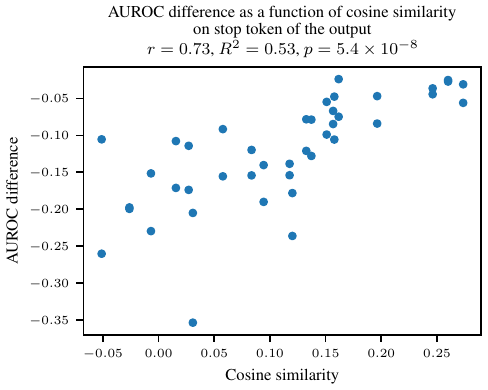}
    \caption{AUROC difference to probe trained on the right dataset as a function of cosine similarity between probes. Results are averaged over 5 runs.}
\end{figure}

\begin{figure}[h]
    \centering
    \includegraphics[width=\columnwidth]{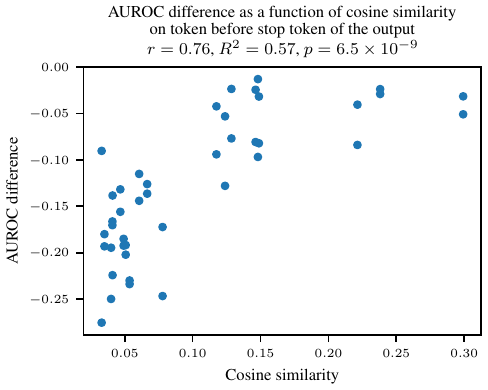}
    \caption{AUROC difference to probe trained on the right dataset as a function of cosine similarity between probes. Results are averaged over 5 runs.}
\end{figure}

\begin{figure}[h]
    \centering
    \includegraphics[width=\columnwidth]{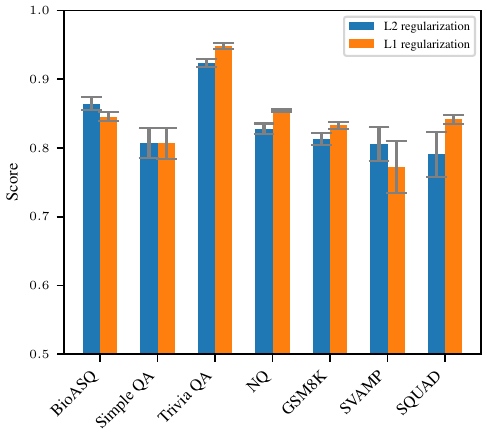}
    \caption{AUROC of linear probes at the stop token of the output, trained with either L1 or L2 regularization.}
\end{figure}

\begin{figure}[h]
    \centering
    \includegraphics[width=\columnwidth]{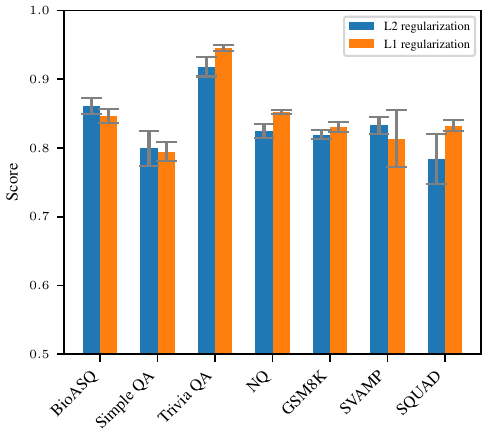}
    \caption{AUROC of linear probes at the token before the stop token of the output, trained with either L1 or L2 regularization.}
\end{figure}

\begin{figure}[h]
    \centering
    \includegraphics[width=\columnwidth]{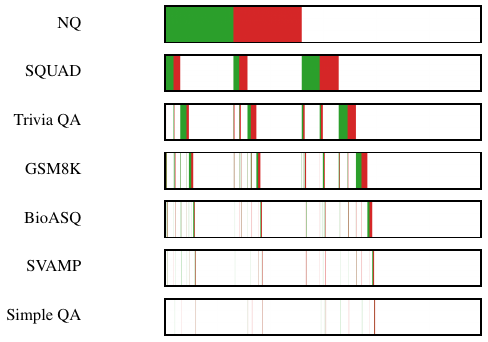}
    \caption{Signed support of sparse probes trained on different datasets at the stop token of the output, using L1 regularization at layer 24. Each row represents one probe trained on the corresponding dataset (y-axis labels). The x-axis shows the 3584 dimensions of the hidden state vector. Green indicates positive coefficients, red indicates negative coefficients, and white indicates zero coefficients (sparsity). Dimensions are sorted by sparsity level across all datasets, with the least sparse dimensions on the left and the most sparse on the right.}
\end{figure}

\begin{figure}[h]
    \centering
    \includegraphics[width=\columnwidth]{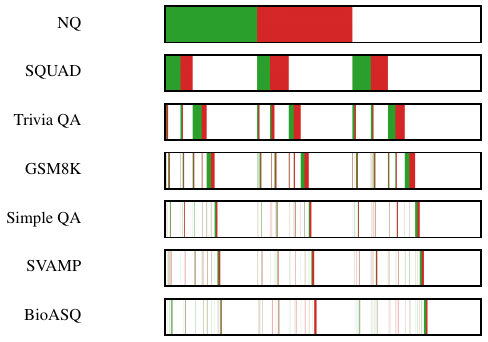}
    \caption{Signed support of sparse probes trained on different datasets at the token before the stop token of the output, using L1 regularization at layer 24. Each row represents one probe trained on the corresponding dataset (y-axis labels). The x-axis shows the 3584 dimensions of the hidden state vector. Green indicates positive coefficients, red indicates negative coefficients, and white indicates zero coefficients (sparsity). Dimensions are sorted by sparsity level across all datasets, with the least sparse dimensions on the left and the most sparse on the right.}
\end{figure}

\begin{figure}[h]
    \centering
    \includegraphics[width=\columnwidth]{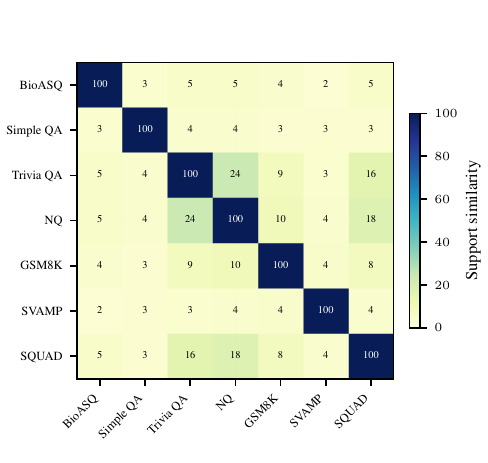}
    \caption{Support overlap between sparse probes trained on different datasets. Darker colors indicate higher overlap percentages. Task pairs with $>30\%$ overlap (TriviaQA, NQ, SimpleQA) correspond to successful cross-task generalization, while most pairs show $<15\%$ overlap, explaining generalization failure.}
\end{figure}

\begin{figure}[h]
    \centering
    \includegraphics[width=\columnwidth]{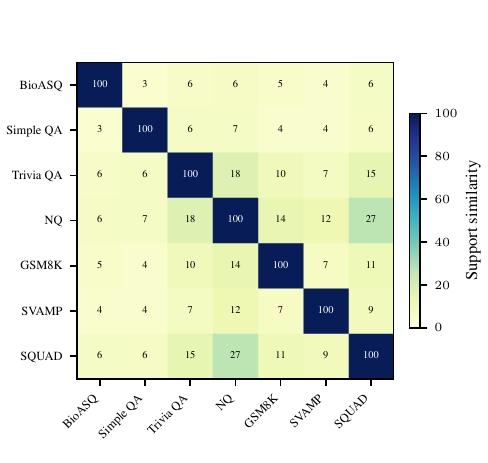}
    \caption{Support overlap between sparse probes trained on different datasets. Darker colors indicate higher overlap percentages. Task pairs with $>30\%$ overlap (TriviaQA, NQ, SimpleQA) correspond to successful cross-task generalization, while most pairs show $<15\%$ overlap, explaining generalization failure.}
\end{figure}

\begin{figure}[h]
    \centering
    \includegraphics[width=\columnwidth]{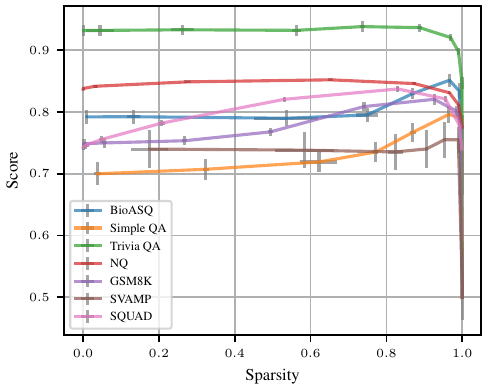}
    \caption{AUROC of probes trained using L1 regularisation as a function of the sparsity level on the stop token of the output.}
\end{figure}

\begin{figure}[h]
    \centering
    \includegraphics[width=\columnwidth]{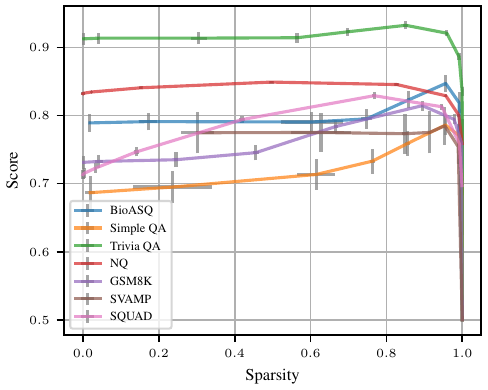}
    \caption{AUROC of probes trained using L1 regularisation as a function of the sparsity level on the token before the stop token of the output.}
\end{figure}

\begin{figure}[h]
    \centering
    \begin{subfigure}[t]{\columnwidth}
    \includegraphics[width=\textwidth]{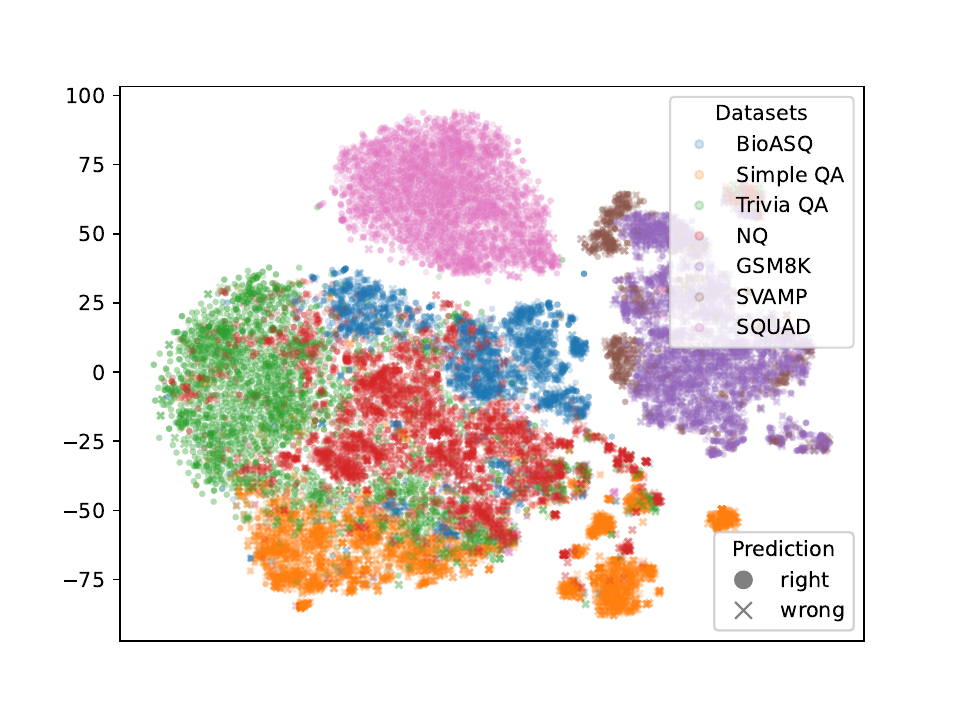}
    \end{subfigure}
    \caption{t-SNE plots of the hidden space at layer 24 at the stop token of the output.}
\end{figure}

\begin{figure}[h]
    \centering
    \begin{subfigure}[t]{\columnwidth}
    \includegraphics[width=\textwidth]{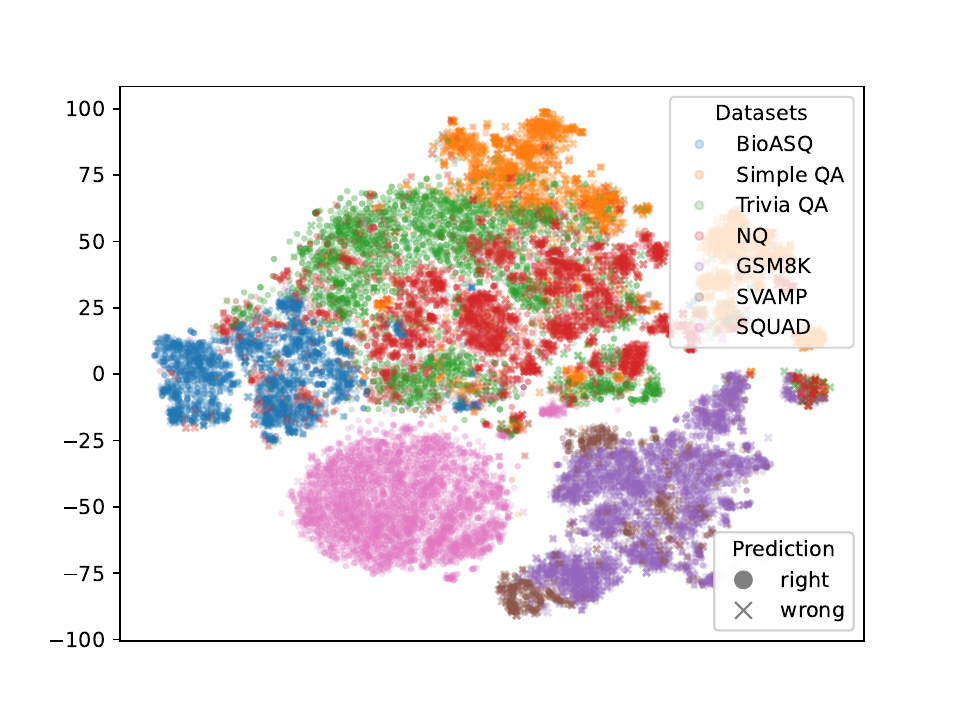}
    \end{subfigure}
    \caption{t-SNE plots of the hidden space at layer 24 at the token before stop token of the output.}
\end{figure}

\begin{figure*}[!h]
    \centering
    \includegraphics[width=\linewidth]{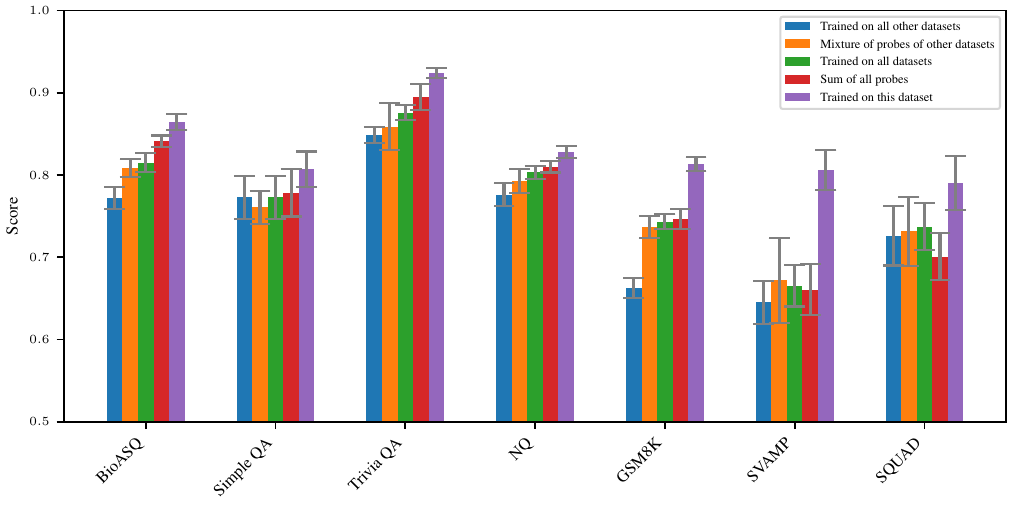}
    \caption{AUROC of linear probes at the stop token of the output in the multi-task setting, using L2 regularization.}
\end{figure*}

\begin{figure*}[!h]
    \centering
    \includegraphics[width=\linewidth]{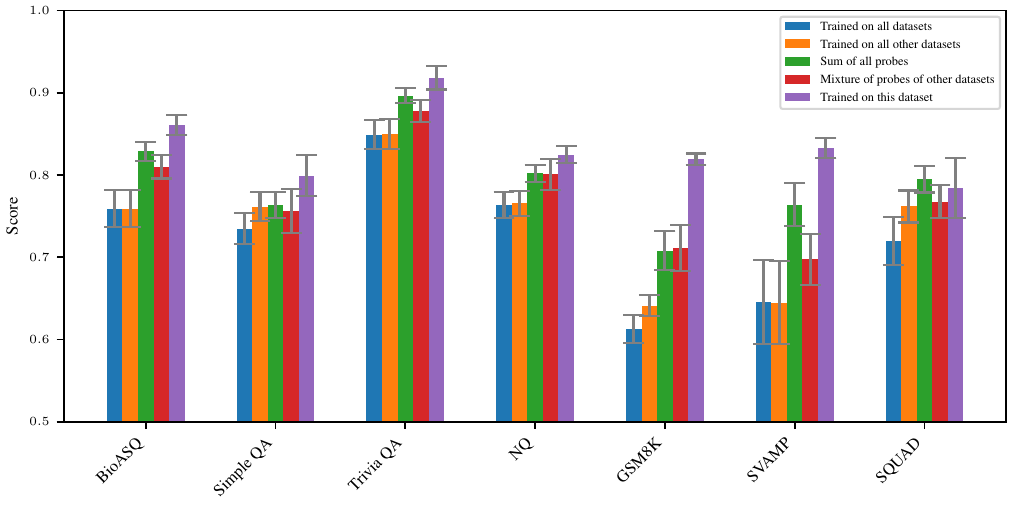}
    \caption{AUROC of linear probes at the token before the stop token of the output in the multi-task setting, using L2 regularization.}
\end{figure*}

\FloatBarrier

\section{Phi-4 Mini Instruct}
\subsection{Layer 32}

\begin{figure}[h]
    \centering
    \begin{subfigure}[t]{\linewidth}
    \includegraphics[width=\linewidth,trim={0.cm 0.cm 0cm 0.2cm},clip]{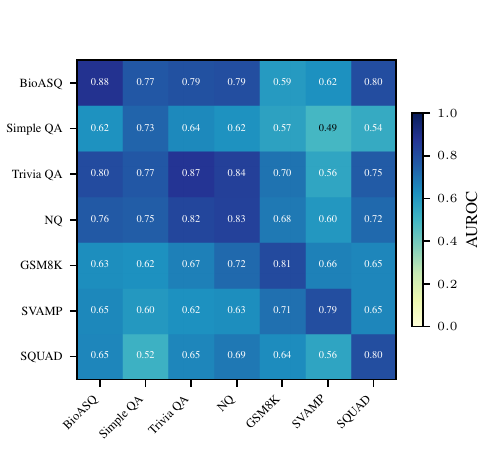}
    \end{subfigure}
    \begin{subfigure}[t]{\linewidth}
    \includegraphics[width=\linewidth,trim={0.cm 0.cm 0.cm 0.2cm},clip]{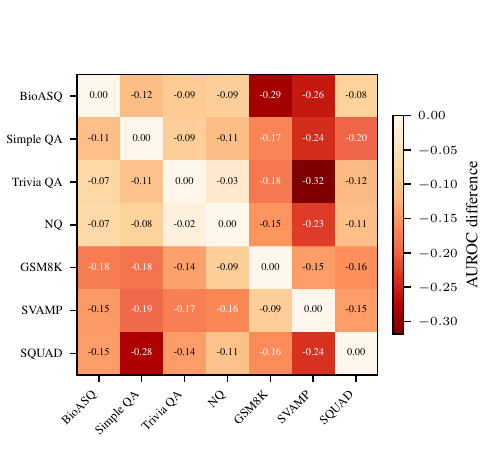}
    \end{subfigure}
    \caption{AUROC of probes trained on different tasks on the stop token of the output on last layer. Rows correspond to evaluation tasks while columns correspond to training tasks. The second plot represents the difference between the probe trained on this task and probes trained on the other datasets. Results are averaged over 5 runs.}
\end{figure}

\begin{figure}[h]
    \centering
    \begin{subfigure}[t]{\linewidth}
    \includegraphics[width=\linewidth,trim={0.cm 0.cm 0cm 0.2cm},clip]{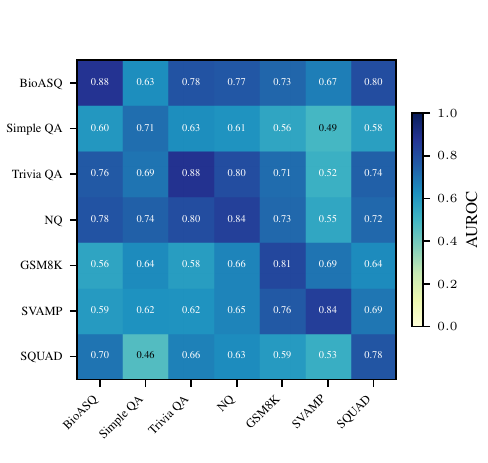}
    \end{subfigure}
    \begin{subfigure}[t]{\linewidth}
    \includegraphics[width=\linewidth,trim={0.cm 0.cm 0.cm 0.2cm},clip]{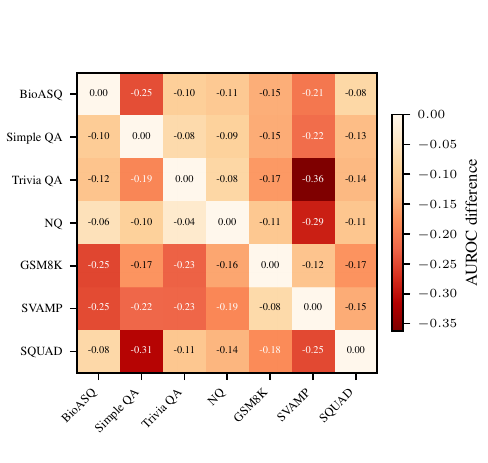}
    \end{subfigure}
    \caption{AUROC of probes trained on different tasks on the token before the stop token of the output on last layer. Rows correspond to evaluation tasks while columns correspond to training tasks. The second plot represents the difference between the probe trained on this task and probes trained on the other datasets. Results are averaged over 5 runs.}
\end{figure}

\begin{figure}[h]
    \centering
    \includegraphics[width=\columnwidth]{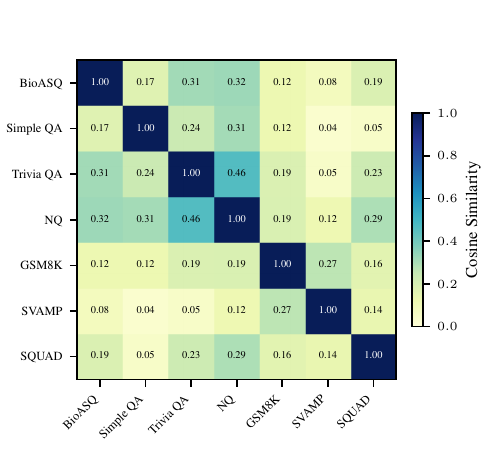}
    \caption{Cosine similarity between probes trained on different datasets using L2 regularization. Results are averaged over 5 runs.
    }
\end{figure}

\begin{figure}[h]
    \centering
    \includegraphics[width=\columnwidth]{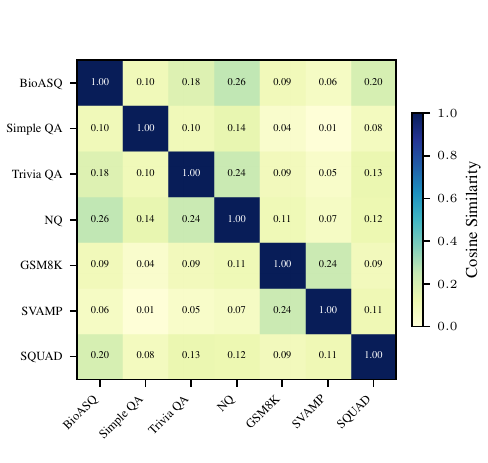}
    \caption{Cosine similarity between probes trained on different datasets using L2 regularization. Results are averaged over 5 runs.
    }
\end{figure}

\begin{figure}[h]
    \centering
    \includegraphics[width=\columnwidth]{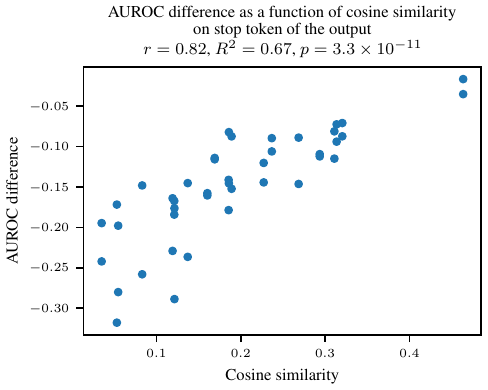}
    \caption{AUROC difference to probe trained on the right dataset as a function of cosine similarity between probes. Results are averaged over 5 runs.}
\end{figure}

\begin{figure}[h]
    \centering
    \includegraphics[width=\columnwidth]{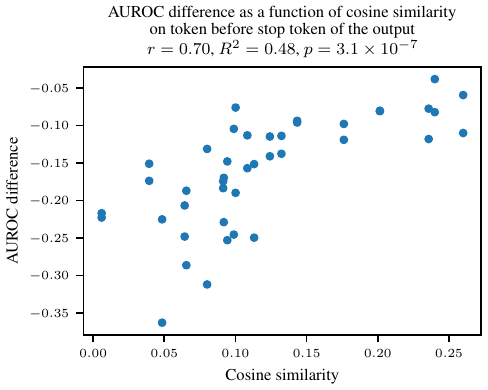}
    \caption{AUROC difference to probe trained on the right dataset as a function of cosine similarity between probes. Results are averaged over 5 runs.}
\end{figure}

\begin{figure}[h]
    \centering
    \includegraphics[width=\columnwidth]{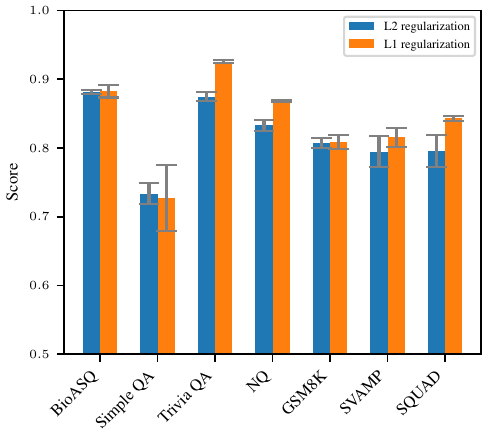}
    \caption{AUROC of linear probes at the stop token of the output, trained with either L1 or L2 regularization.}
\end{figure}

\begin{figure}[h]
    \centering
    \includegraphics[width=\columnwidth]{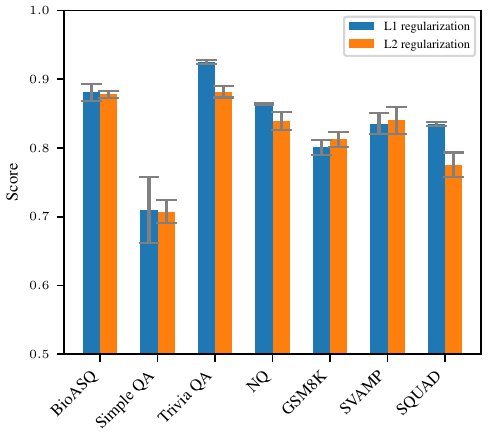}
    \caption{AUROC of linear probes at the token before the stop token of the output, trained with either L1 or L2 regularization.}
\end{figure}

\begin{figure}[h]
    \centering
    \includegraphics[width=\columnwidth]{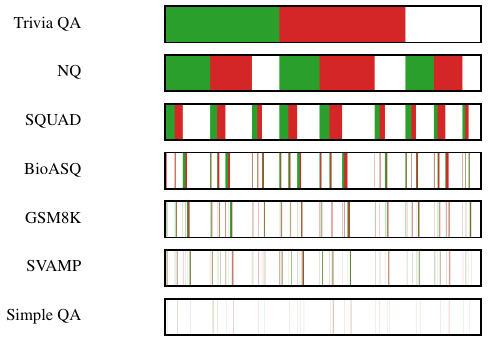}
    \caption{Signed support of sparse probes trained on different datasets at the stop token of the output, using L1 regularization at layer 32. Each row represents one probe trained on the corresponding dataset (y-axis labels). The x-axis shows the 3584 dimensions of the hidden state vector. Green indicates positive coefficients, red indicates negative coefficients, and white indicates zero coefficients (sparsity). Dimensions are sorted by sparsity level across all datasets, with the least sparse dimensions on the left and the most sparse on the right.}
\end{figure}

\begin{figure}[h]
    \centering
    \includegraphics[width=\columnwidth]{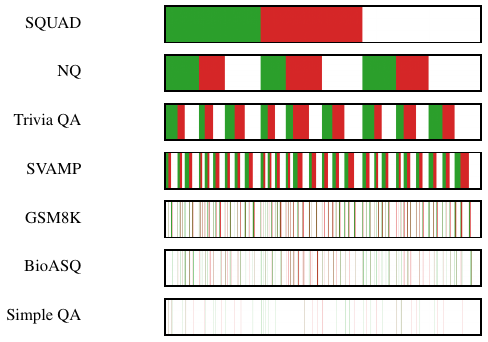}
    \caption{Signed support of sparse probes trained on different datasets at the token before the stop token of the output, using L1 regularization at layer 32. Each row represents one probe trained on the corresponding dataset (y-axis labels). The x-axis shows the 3584 dimensions of the hidden state vector. Green indicates positive coefficients, red indicates negative coefficients, and white indicates zero coefficients (sparsity). Dimensions are sorted by sparsity level across all datasets, with the least sparse dimensions on the left and the most sparse on the right.}
\end{figure}

\begin{figure}[h]
    \centering
    \includegraphics[width=\columnwidth]{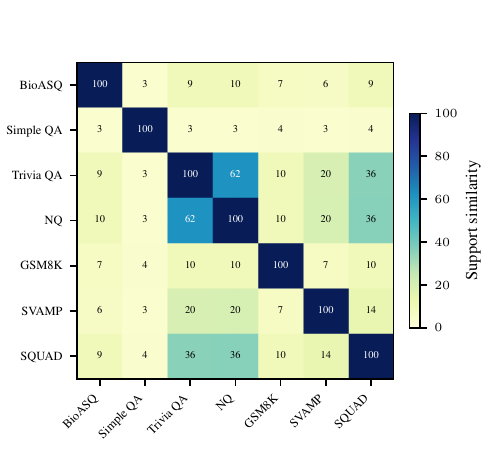}
    \caption{Support overlap between sparse probes trained on different datasets. Darker colors indicate higher overlap percentages. Task pairs with $>30\%$ overlap (TriviaQA, NQ, SimpleQA) correspond to successful cross-task generalization, while most pairs show $<15\%$ overlap, explaining generalization failure.}
\end{figure}

\begin{figure}[h]
    \centering
    \includegraphics[width=\columnwidth]{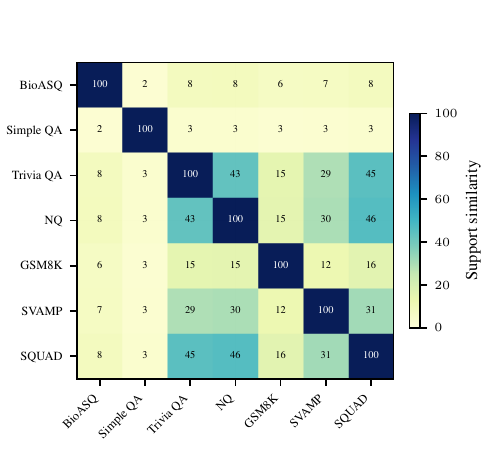}
    \caption{Support overlap between sparse probes trained on different datasets. Darker colors indicate higher overlap percentages. Task pairs with $>30\%$ overlap (TriviaQA, NQ, SimpleQA) correspond to successful cross-task generalization, while most pairs show $<15\%$ overlap, explaining generalization failure.}
\end{figure}

\begin{figure}[h]
    \centering
    \includegraphics[width=\columnwidth]{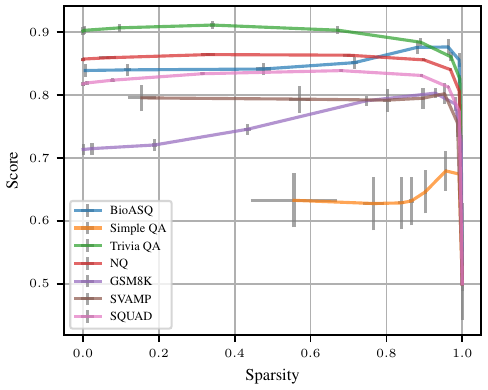}
    \caption{AUROC of probes trained using L1 regularisation as a function of the sparsity level on the stop token of the output.}
\end{figure}

\begin{figure}[h]
    \centering
    \includegraphics[width=\columnwidth]{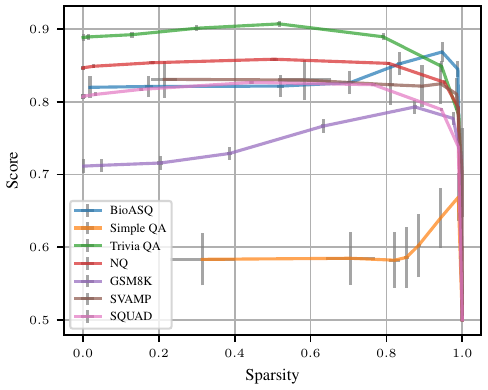}
    \caption{AUROC of probes trained using L1 regularisation as a function of the sparsity level on the token before the stop token of the output.}
\end{figure}

\begin{figure}[h]
    \centering
    \begin{subfigure}[t]{\columnwidth}
    \includegraphics[width=\textwidth]{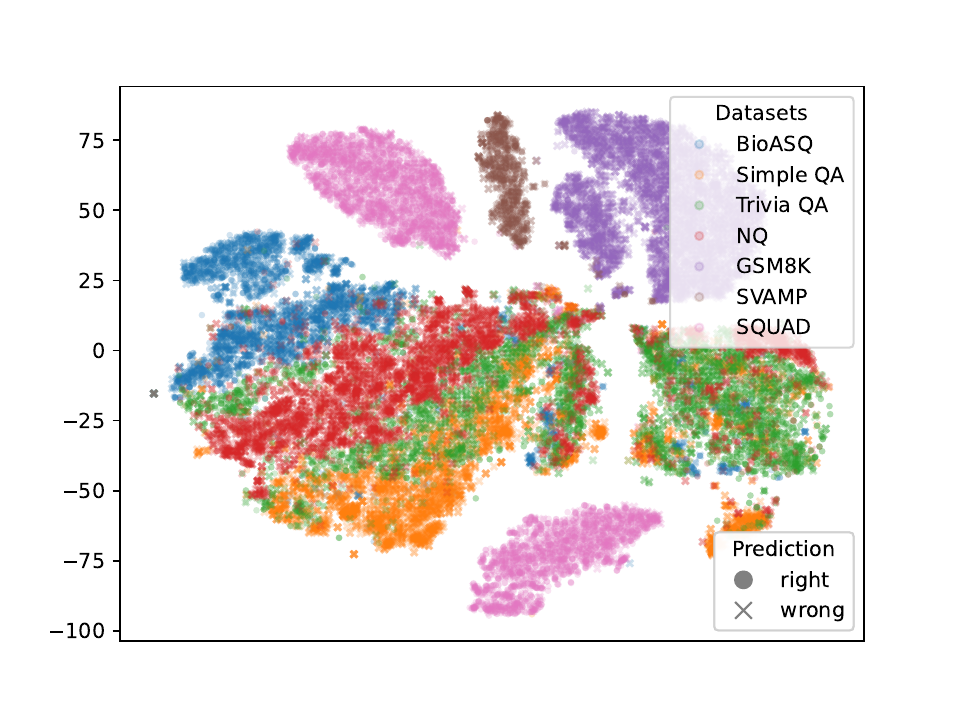}
    \end{subfigure}
    \caption{t-SNE plots of the hidden space at layer 32 at the stop token of the output.}
\end{figure}

\begin{figure}[h]
    \centering
    \begin{subfigure}[t]{\columnwidth}
    \includegraphics[width=\textwidth]{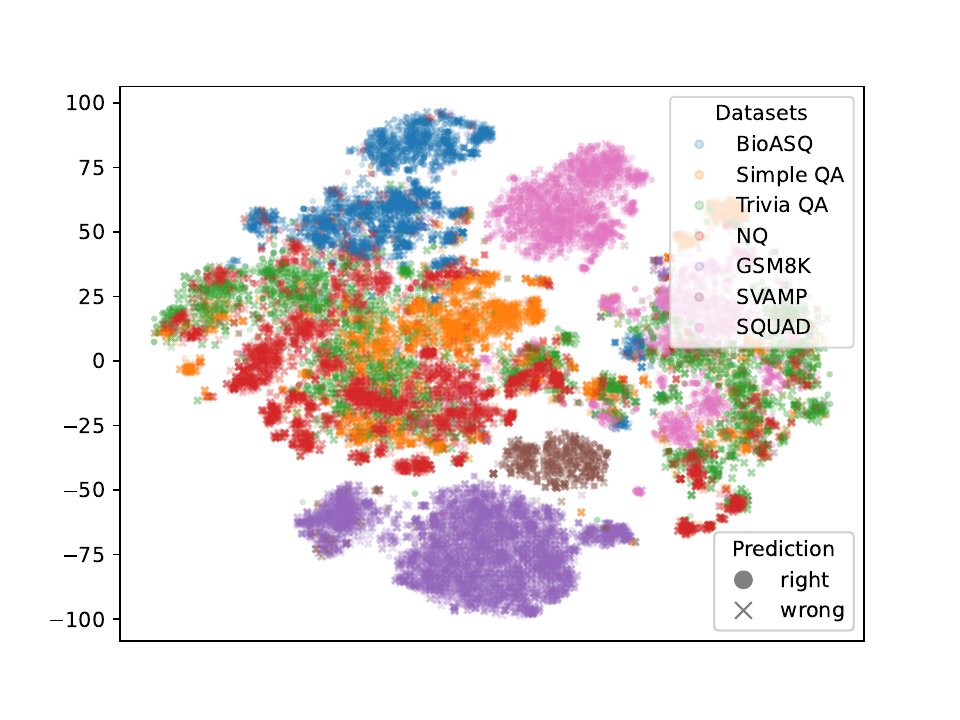}
    \end{subfigure}
    \caption{t-SNE plots of the hidden space at layer 32 at the token before stop token of the output.}
\end{figure}

\begin{figure*}[!h]
    \centering
    \includegraphics[width=\linewidth]{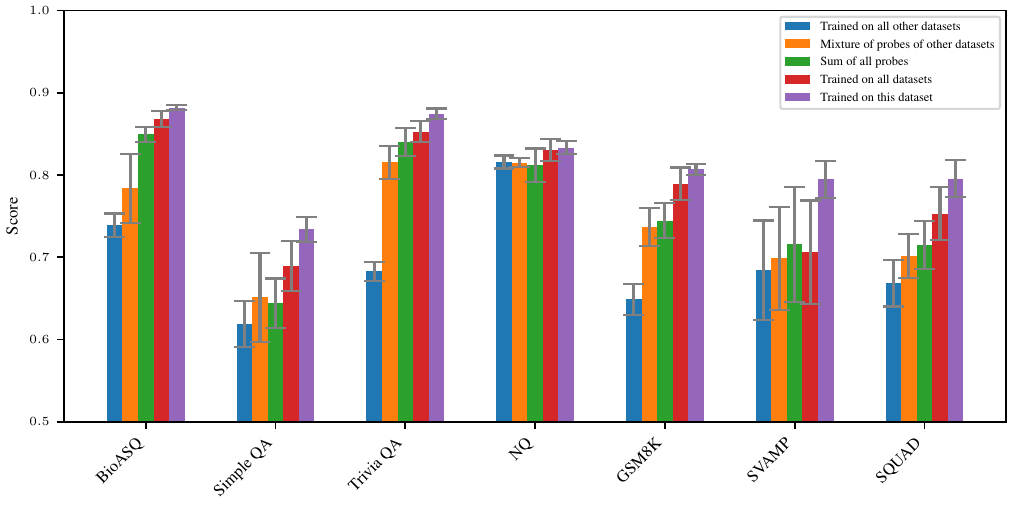}
    \caption{AUROC of linear probes at the stop token of the output in the multi-task setting, using L2 regularization.}
\end{figure*}

\begin{figure*}[!h]
    \centering
    \includegraphics[width=\linewidth]{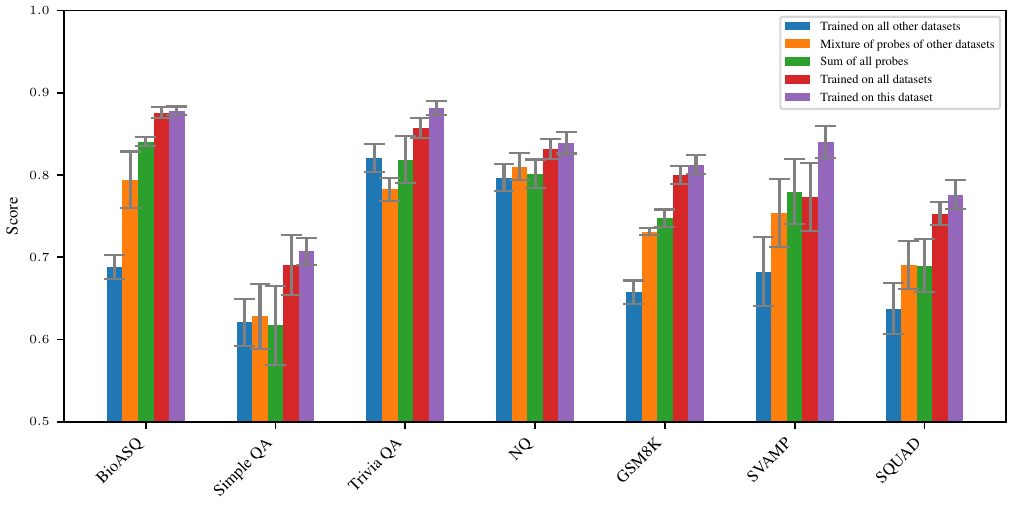}
    \caption{AUROC of linear probes at the token before the stop token of the output in the multi-task setting, using L2 regularization.}
\end{figure*}

\FloatBarrier

\subsection{Layer 24}

\begin{figure}[h]
    \centering
    \begin{subfigure}[t]{\linewidth}
    \includegraphics[width=\linewidth,trim={0.cm 0.cm 0cm 0.2cm},clip]{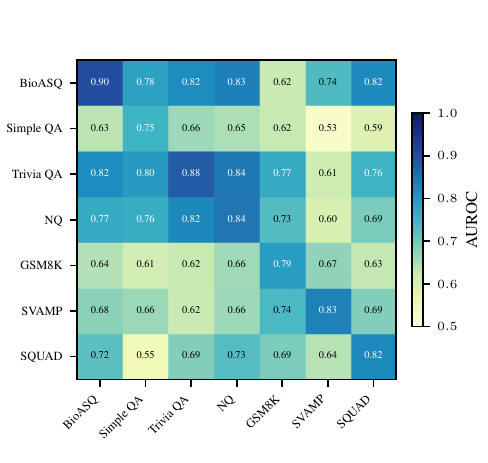}
    \end{subfigure}
    \begin{subfigure}[t]{\linewidth}
    \includegraphics[width=\linewidth,trim={0.cm 0.cm 0.cm 0.2cm},clip]{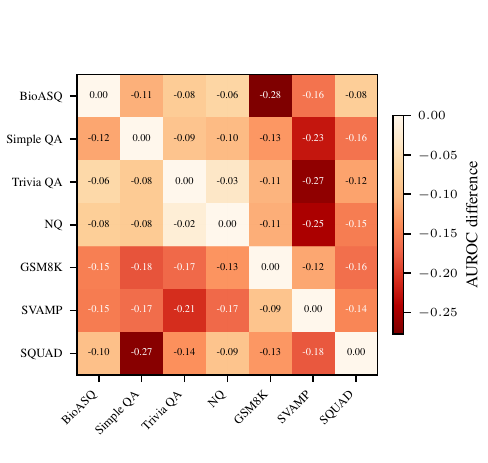}
    \end{subfigure}
    \caption{AUROC of probes trained on different tasks on the stop token of the output on last layer. Rows correspond to evaluation tasks while columns correspond to training tasks. The second plot represents the difference between the probe trained on this task and probes trained on the other datasets. Results are averaged over 5 runs.}
\end{figure}

\begin{figure}[h]
    \centering
    \begin{subfigure}[t]{\linewidth}
    \includegraphics[width=\linewidth,trim={0.cm 0.cm 0cm 0.2cm},clip]{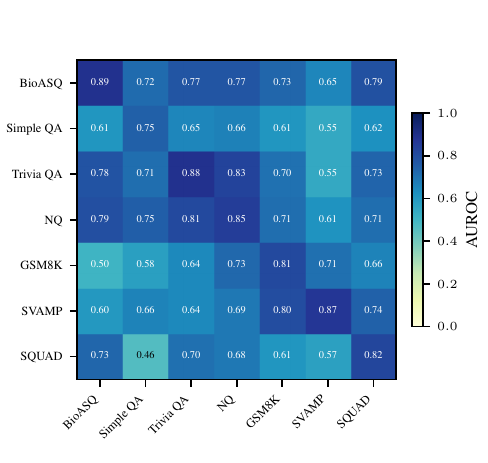}
    \end{subfigure}
    \begin{subfigure}[t]{\linewidth}
    \includegraphics[width=\linewidth,trim={0.cm 0.cm 0.cm 0.2cm},clip]{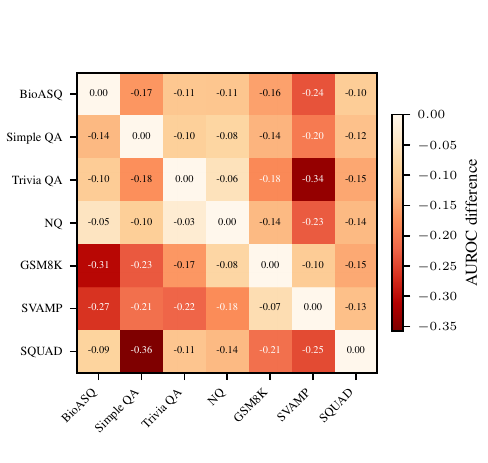}
    \end{subfigure}
    \caption{AUROC of probes trained on different tasks on the token before the stop token of the output on last layer. Rows correspond to evaluation tasks while columns correspond to training tasks. The second plot represents the difference between the probe trained on this task and probes trained on the other datasets. Results are averaged over 5 runs.}
\end{figure}

\begin{figure}[h]
    \centering
    \includegraphics[width=\columnwidth]{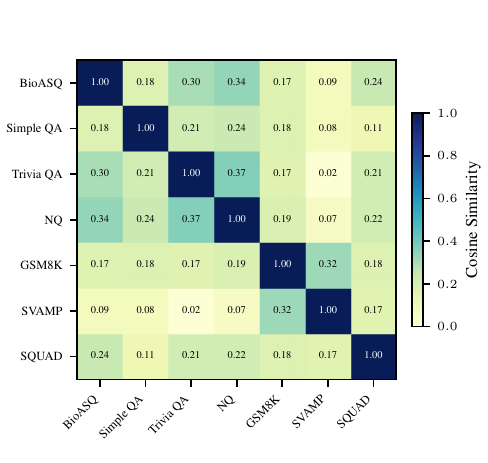}
    \caption{Cosine similarity between probes trained on different datasets using L2 regularization. Results are averaged over 5 runs.
    }
\end{figure}

\begin{figure}[h]
    \centering
    \includegraphics[width=\columnwidth]{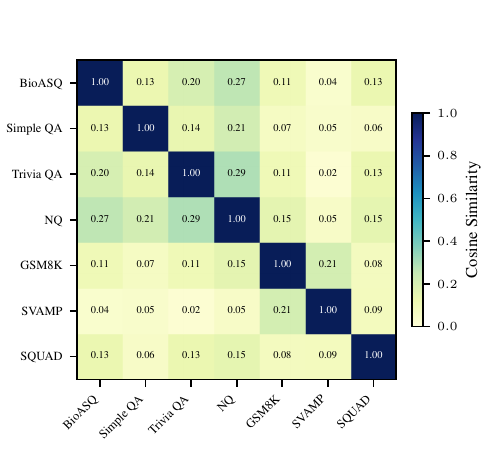}
    \caption{Cosine similarity between probes trained on different datasets using L2 regularization. Results are averaged over 5 runs.
    }
\end{figure}

\begin{figure}[h]
    \centering
    \includegraphics[width=\columnwidth]{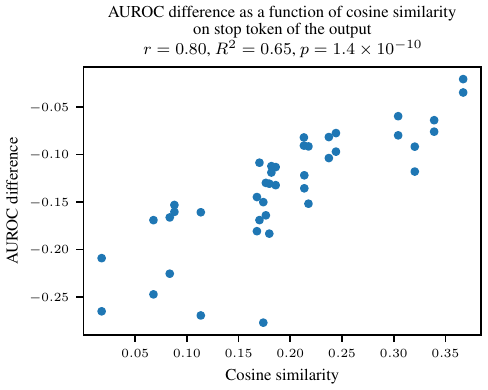}
    \caption{AUROC difference to probe trained on the right dataset as a function of cosine similarity between probes. Results are averaged over 5 runs.}
\end{figure}

\begin{figure}[h]
    \centering
    \includegraphics[width=\columnwidth]{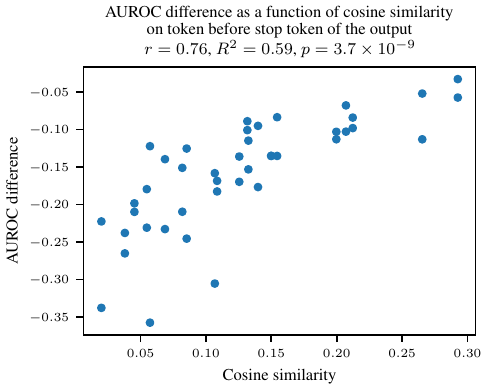}
    \caption{AUROC difference to probe trained on the right dataset as a function of cosine similarity between probes. Results are averaged over 5 runs.}
\end{figure}

\begin{figure}[h]
    \centering
    \includegraphics[width=\columnwidth]{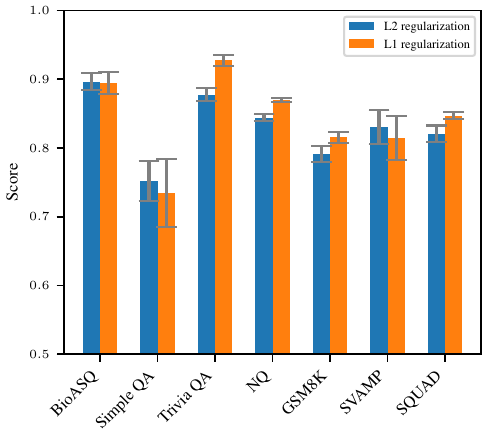}
    \caption{AUROC of linear probes at the stop token of the output, trained with either L1 or L2 regularization.}
\end{figure}

\begin{figure}[h]
    \centering
    \includegraphics[width=\columnwidth]{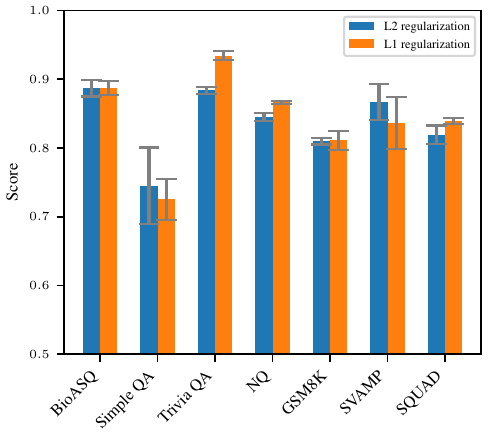}
    \caption{AUROC of linear probes at the token before the stop token of the output, trained with either L1 or L2 regularization.}
\end{figure}

\begin{figure}[h]
    \centering
    \includegraphics[width=\columnwidth]{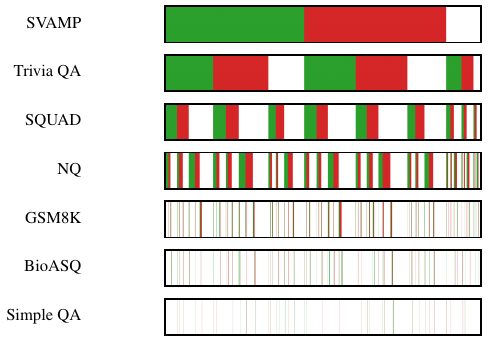}
    \caption{Signed support of sparse probes trained on different datasets at the stop token of the output, using L1 regularization at layer 24. Each row represents one probe trained on the corresponding dataset (y-axis labels). The x-axis shows the 3584 dimensions of the hidden state vector. Green indicates positive coefficients, red indicates negative coefficients, and white indicates zero coefficients (sparsity). Dimensions are sorted by sparsity level across all datasets, with the least sparse dimensions on the left and the most sparse on the right.}
\end{figure}

\begin{figure}[h]
    \centering
    \includegraphics[width=\columnwidth]{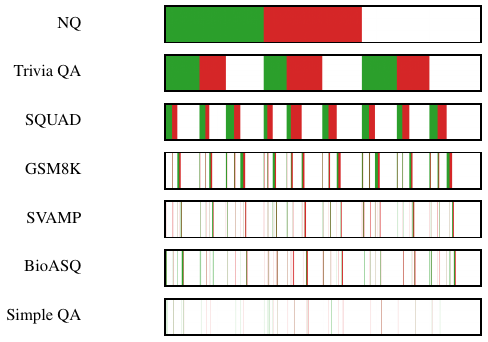}
    \caption{Signed support of sparse probes trained on different datasets at the token before the stop token of the output, using L1 regularization at layer 24. Each row represents one probe trained on the corresponding dataset (y-axis labels). The x-axis shows the 3584 dimensions of the hidden state vector. Green indicates positive coefficients, red indicates negative coefficients, and white indicates zero coefficients (sparsity). Dimensions are sorted by sparsity level across all datasets, with the least sparse dimensions on the left and the most sparse on the right.}
\end{figure}

\begin{figure}[h]
    \centering
    \includegraphics[width=\columnwidth]{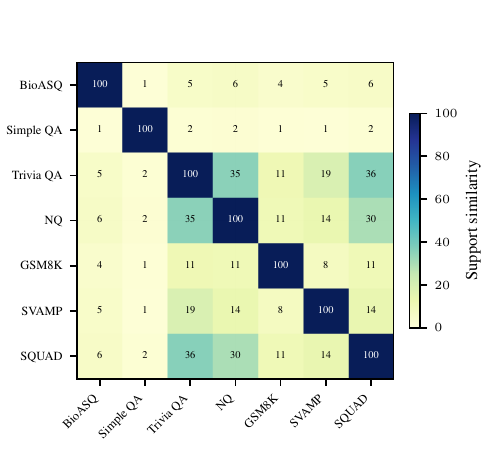}
    \caption{Support overlap between sparse probes trained on different datasets. Darker colors indicate higher overlap percentages. Task pairs with $>30\%$ overlap (TriviaQA, NQ, SimpleQA) correspond to successful cross-task generalization, while most pairs show $<15\%$ overlap, explaining generalization failure.}
\end{figure}

\begin{figure}[h]
    \centering
    \includegraphics[width=\columnwidth]{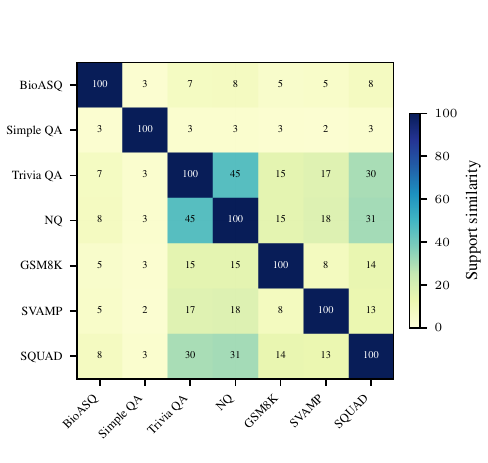}
    \caption{Support overlap between sparse probes trained on different datasets. Darker colors indicate higher overlap percentages. Task pairs with $>30\%$ overlap (TriviaQA, NQ, SimpleQA) correspond to successful cross-task generalization, while most pairs show $<15\%$ overlap, explaining generalization failure.}
\end{figure}

\begin{figure}[h]
    \centering
    \includegraphics[width=\columnwidth]{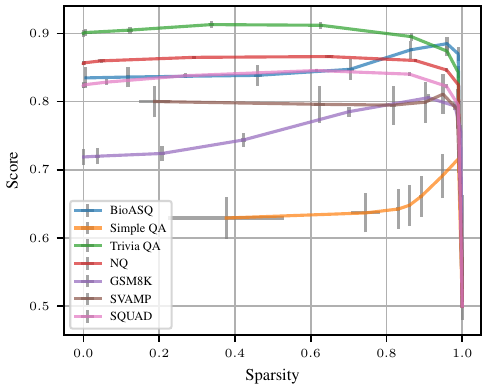}
    \caption{AUROC of probes trained using L1 regularisation as a function of the sparsity level on the stop token of the output.}
\end{figure}

\begin{figure}[h]
    \centering
    \includegraphics[width=\columnwidth]{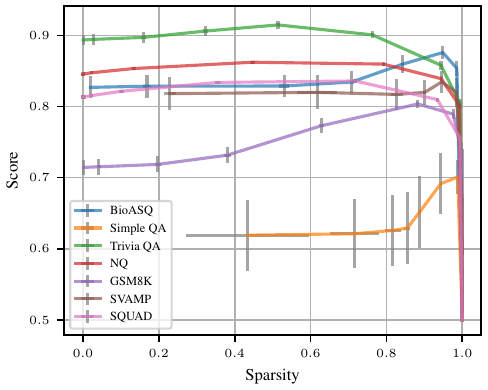}
    \caption{AUROC of probes trained using L1 regularisation as a function of the sparsity level on the token before the stop token of the output.}
\end{figure}

\begin{figure}[h]
    \centering
    \begin{subfigure}[t]{\columnwidth}
    \includegraphics[width=\textwidth]{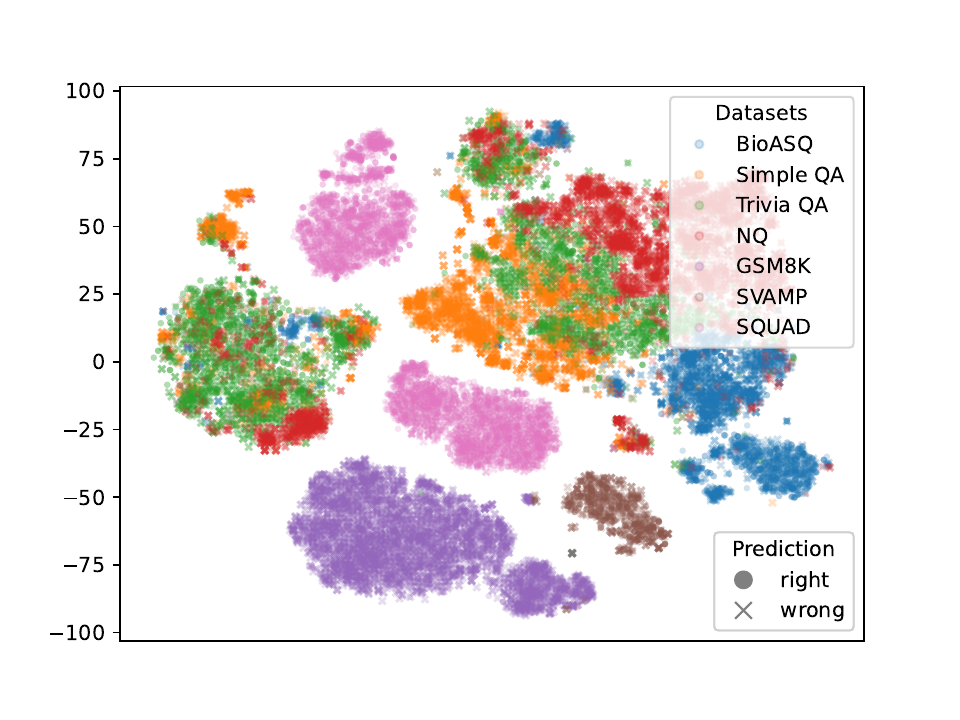}
    \end{subfigure}
    \caption{t-SNE plots of the hidden space at layer 24 at the stop token of the output.}
\end{figure}

\begin{figure}[h]
    \centering
    \begin{subfigure}[t]{\columnwidth}
    \includegraphics[width=\textwidth]{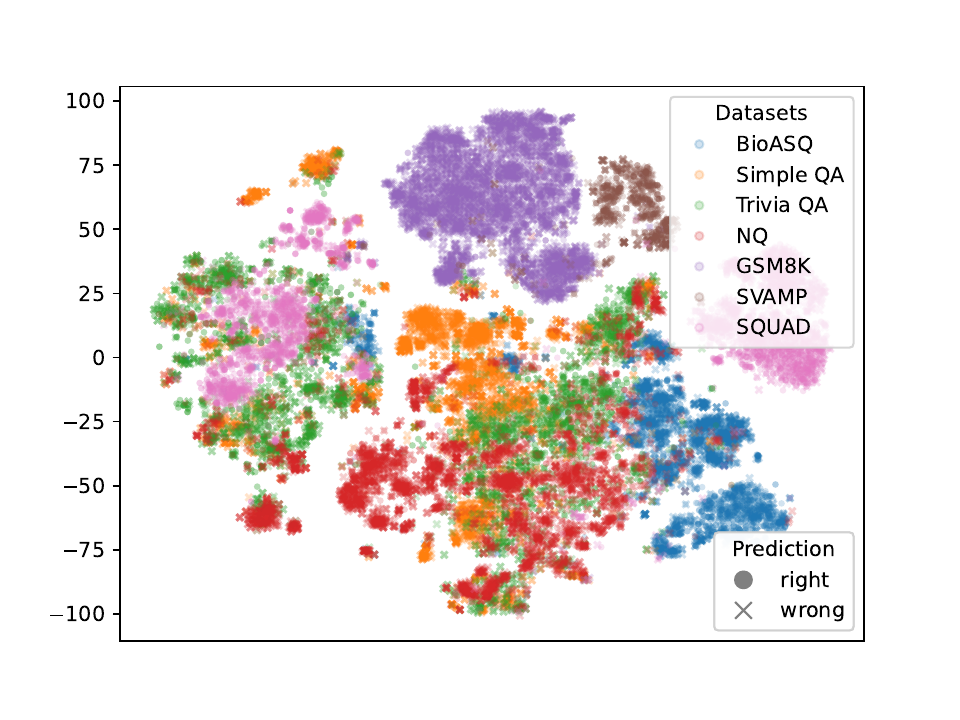}
    \end{subfigure}
    \caption{t-SNE plots of the hidden space at layer 24 at the token before stop token of the output.}
\end{figure}

\begin{figure*}[!h]
    \centering
    \includegraphics[width=\linewidth]{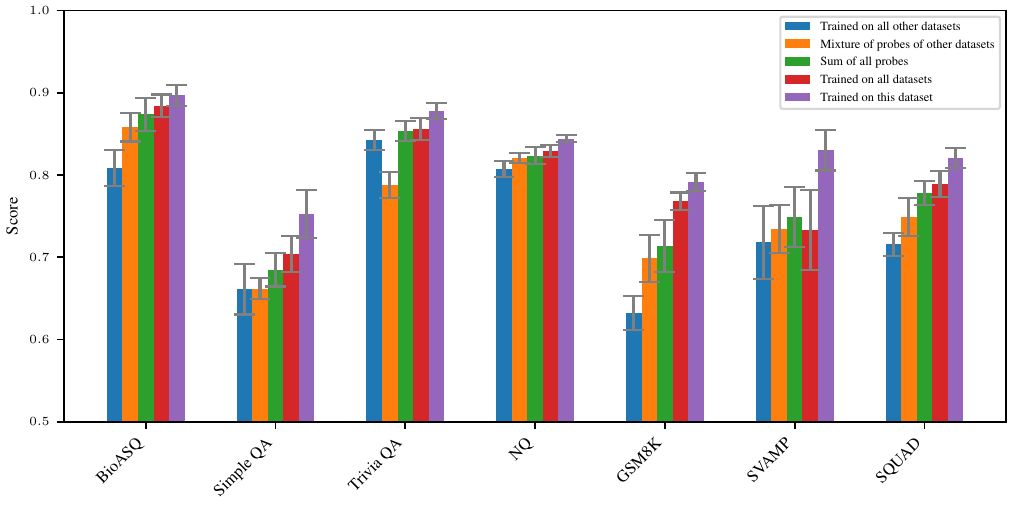}
    \caption{AUROC of linear probes at the stop token of the output in the multi-task setting, using L2 regularization.}
\end{figure*}

\begin{figure*}[!h]
    \centering
    \includegraphics[width=\linewidth]{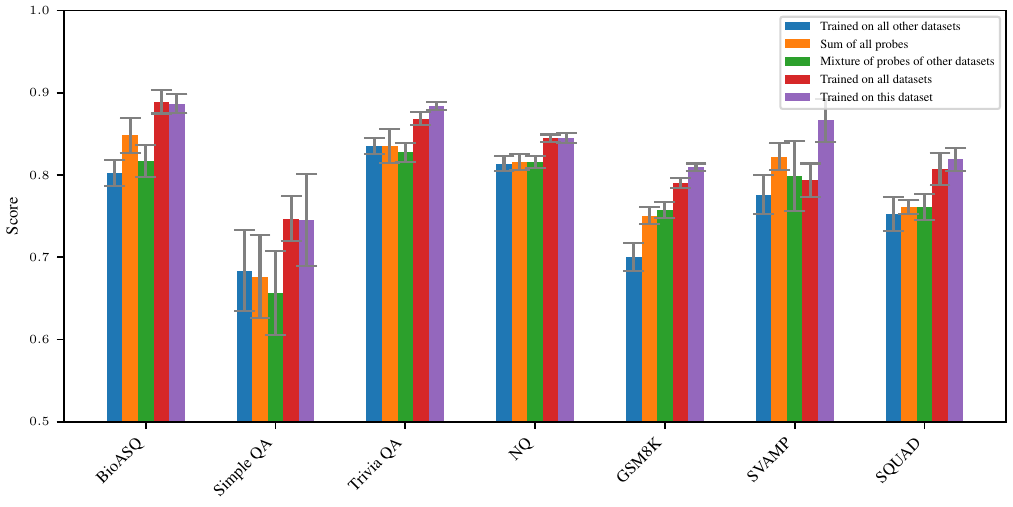}
    \caption{AUROC of linear probes at the token before the stop token of the output in the multi-task setting, using L2 regularization.}
\end{figure*}

\FloatBarrier

\end{document}